\documentclass[10pt,journal,compsoc]{IEEEtran}

\ifCLASSOPTIONcompsoc
 \usepackage[nocompress]{cite}
\else
 \usepackage{cite}
\fi

\usepackage{array}
\usepackage{amsmath,amssymb,amsfonts} 
\usepackage{adjustbox}
\usepackage{bm,bbm} %
\usepackage{caption}
\usepackage{soul}
\usepackage{color}
\usepackage{xspace}
\usepackage{float}
\usepackage[edges]{forest}
\usepackage{graphicx}

\usepackage{multirow,multicol}
\usepackage{overpic}
\usepackage{subfigure}
\usepackage{tikz}
\usepackage{pgfplots}
\pgfplotsset{compat=newest}
\usetikzlibrary{external}
\usetikzlibrary{trees,positioning,shapes,shadows,arrows.meta}

\usepackage{threeparttable}

\usepackage{hyperref}
\hypersetup{colorlinks,linkcolor={blue},citecolor={blue},urlcolor={red}}

\usepackage[capitalize]{cleveref}

\crefname{section}{Sec.}{Secs.}
\crefname{section}{Section}{Sections}
\crefname{table}{Table}{Tables}
\crefname{table}{Tab.}{Tabs.}
\usepackage[switch]{lineno}
\usepackage[normalem]{ulem}

\usepackage{pifont}
\newcommand{\cmark}{\ding{51}}%
\newcommand{\xmark}{\ding{55}}%

\usepackage{pdfrender}
\newcommand*{\boldcheckmark}{%
 \textpdfrender{
 TextRenderingMode=FillStroke,
 LineWidth=.5pt, %
 }{\checkmark}%
}

\let\oldput\put
\def\put(#1,#2)#3{%
 \oldput(#1,#2){\scriptsize #3}%
}

\makeatletter
\DeclareRobustCommand\onedot{\futurelet\@let@token\@onedot}
\def\@onedot{\ifx\@let@token.\else.\null\fi\xspace}

\def\eg{\emph{e.g}\onedot} 
\def\ie{\emph{i.e}\onedot} 
\def\etal{\emph{et al}\onedot}

\DeclareMathOperator*{\argmax}{arg\,max}

\DeclareMathOperator*{\assignment}{\mathcal{A}}
\definecolor{apricot}{rgb}{0.98, 0.81, 0.69}
\definecolor{asparagus}{rgb}{0.53, 0.66, 0.42}
\definecolor{palerobineggblue}{rgb}{0.59, 0.87, 0.82}
\definecolor{olivine}{rgb}{0.6, 0.73, 0.45}
\definecolor{oldlace}{rgb}{0.99, 0.96, 0.9}
\definecolor{airforceblue}{rgb}{0.36, 0.54, 0.66}
\definecolor{amaranth}{rgb}{0.9, 0.17, 0.31}
\definecolor{babypink}{rgb}{0.96, 0.76, 0.76}
\definecolor{cadetblue}{rgb}{0.37, 0.62, 0.63}
\definecolor{cambridgeblue}{rgb}{0.64, 0.76, 0.68}
\definecolor{carolinablue}{rgb}{0.6, 0.73, 0.89}
\definecolor{darkcyan}{rgb}{0.0, 0.55, 0.55}
\definecolor{etonblue}{rgb}{0.59, 0.78, 0.64}
\definecolor{indianyellow}{rgb}{0.89, 0.66, 0.34}
\definecolor{lilac}{rgb}{0.78, 0.64, 0.78}
\definecolor{pastelred}{rgb}{1.0, 0.41, 0.38}
\definecolor{sandstorm}{rgb}{0.93, 0.84, 0.25}
\definecolor{sunglow}{rgb}{1.0, 0.8, 0.2}
\definecolor{greenff}{rgb}{0.0, 0.5, 0.0}
\definecolor{LightCyan}{rgb}{0.88,1,1}

\newcommand{\etc}{\textit{etc}}
\DeclareMathOperator*{\argmin}{arg\,min}

\begin{document}

\title{Temporal Action Segmentation:\\ 
An Analysis of Modern Techniques}

\author{Guodong~Ding$^*$, %
 Fadime~Sener$^*$, %
 and~Angela~Yao %
\IEEEcompsocitemizethanks{
\IEEEcompsocthanksitem Guodong Ding and Angela Yao are with the School of Computing, National University of Singapore, Singapore (emails: dinggd@comp.nus.edu.sg, ayao@comp.nus.edu.sg). \protect 
\IEEEcompsocthanksitem Fadime Sener is a research scientist at Meta Reality Labs (email: famesener@meta.com).
\IEEEcompsocthanksitem $^*$ indicates equal contribution. \protect}%
\thanks{Manuscript received October 19, 2022}}

\markboth{Journal of \LaTeX\ Class Files,~Vol.~14, No.~8, August~2015}%
{Shell \MakeLowercase{\textit{et al.}}: Bare Advanced Demo of IEEEtran.cls for IEEE Computer Society Journals}

\IEEEtitleabstractindextext{%
\begin{abstract} 
 
Temporal action segmentation (TAS) in videos aims at densely identifying video frames in minutes-long videos with multiple action classes. As a long-range video understanding task, researchers have developed an extended collection of methods and examined their performance using various benchmarks. Despite the rapid growth of TAS techniques in recent years, no systematic survey has been conducted in these sectors. This survey analyzes and summarizes the most significant contributions and trends. In particular, we first examine the task definition, common benchmarks, types of supervision, and prevalent evaluation measures. In addition, we systematically investigate two essential techniques of this topic, \ie, frame representation and temporal modeling, which have been studied extensively in the literature. We then conduct a thorough review of existing TAS works categorized by their levels of supervision and conclude our survey by identifying and emphasizing several research gaps. In addition, we have curated a list of TAS resources, which is available at \url{https://github.com/nus-cvml/awesome-temporal-action-segmentation}.
\end{abstract}

}

\maketitle

\IEEEdisplaynontitleabstractindextext
\IEEEpeerreviewmaketitle

\ifCLASSOPTIONcompsoc
\IEEEraisesectionheading{\section{Introduction}\label{sec:introduction}}
\else
\section{Introduction}\label{sec:introduction}
\fi

\begin{figure}[!tb]
    \centering
    \begin{overpic}[width=0.7\linewidth]{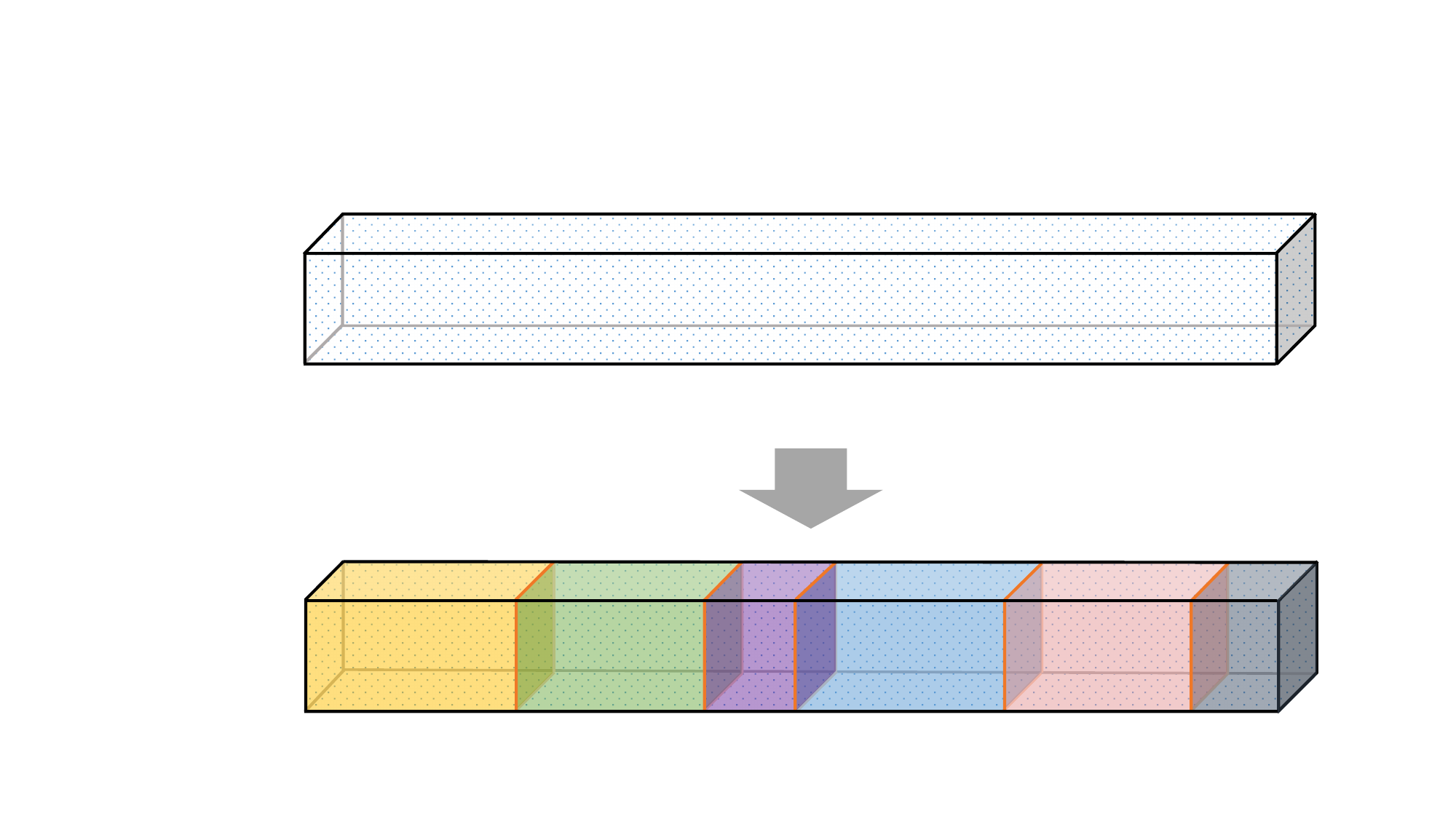}
    \put(27, 36){Untrimmed Procedural Video}
    \put(25,27){Temporal Action Segmentation}
    \put(10,7){Action}
    \put(27,7){Action}
    \put(40,7){Action}
    \put(53,7){Action}
    \put(69,7){Action}
    \put(82,7){Action}
    \put(15,3.6){1}
    \put(32,3.6){2}
    \put(44,3.6){3}
    \put(57,3.6){4}
    \put(74,3.6){5}
    \put(85,3.6){6}
    \end{overpic}
    \caption{A TAS model segments an untrimmed video sequence in the temporal dimension into successive actions. }\label{fig:teaser}
\end{figure}
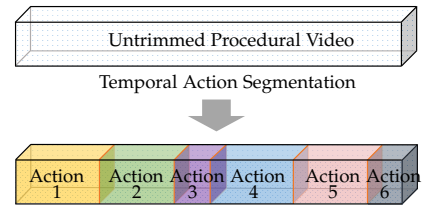

\IEEEPARstart{T}{emporal} action segmentation (TAS) is a video understanding task that segments, in time, a temporally untrimmed video sequence. Each segment is labeled with one of a finite set of pre-defined action labels (see \cref{fig:teaser} for a visual illustration). This task is a 1D temporal analogue to the more established semantic segmentation~\cite{minaee2022image}, replacing pixel-wise semantic labels with frame-wise action labels. Automatically segmenting untrimmed video sequences helps to understand what actions are being performed, when they started, how far they have progressed, %
how the actions transform the environment, %
and what people will do next. It also enables diverse downstream applications, such as assistive technologies, video %
surveillance, and human-robot interactions. This survey introduces the techniques to understand action segmentation and follows with a comprehensive overview of recent %
methods.

In computer vision, \emph{action recognition} is the hallmark task for video understanding. In action recognition, pre-trimmed video clips of a few seconds are classified with single semantic labels. State-of-the-art methods~\cite{feichtenhofer2019slowfast,lin2019tsm,patrick2021keeping} can distinguish hundreds of classes. However, classifying pre-trimmed clips is a highly limiting case as the video feeds of surveillance systems, autonomous vehicles, and other real-world systems occur in streams. The individual actions or events are related and may span well beyond a few seconds. As a result, standard action recognition approaches are not directly applicable. 
 
Action segmentation methods target untrimmed video sequences. %
The videos portray a series of multiple actions, and typically span several minutes. A common \emph{``making coffee''} procedural video may include the following steps: `take cup', `pour coffee', `pour sugar', `pour milk', and `stir coffee'. In the domain of procedural videos, the common terminology to describe the overall procedure is ~\emph{(complex) activity}, whereas the composing steps are~\emph{actions}. Importantly, the steps often adhere to a loose temporal ordering, \ie, permutations of some actions in time (`pour coffee' and `pour milk') and optional actions (`pour sugar'). %

An effective segmentation model should use sequential information to %
determine action boundaries. This leads to two considerations: learning discriminative frame-level representations and modeling temporal and
sequential relationships between actions. Frame-level representations should capture both static and dynamic visual information for discrimination. Furthermore, the sequential dynamics of actions should be well captured. The ordering characteristics of actions raise a fundamental question - how should temporal or sequential relationships be modeled to account for action repetition, duration, and order variations? This survey identifies the aforementioned two aspects as the essential techniques for the temporal action segmentation task and provides respective in-depth analyses.

\textbf{Contributions.}
There are several surveys on human activity understanding in videos, though their focus is primarily on action recognition~\cite{zhang2019comprehensive,kong2022human}, temporal action localization~\cite{xia2020survey,baraka2022weakly}, action anticipation~\cite{trong2017comprehensive,rasouli2020deep}, \etc. To the best of our knowledge, this is the first survey %
of temporal action segmentation. In addition to categorizing existing works, we propose a taxonomy that emphasizes their contributions.

Additionally, this survey analyzes the characteristics of action segmentation datasets. In doing so, it introduces the repetition and order variation scores, two metrics which characterize the temporal dynamics of actions. The analysis shows that the majority of existing datasets are limited in action repetition and order variation. Furthermore, several performance evaluation and comparison settings are distinguished. A standardized evaluation setup is provided for unsupervised segmentation methods, along with a class-based evaluation metric emphasizing the long-tail distribution. Lastly, a handful of intriguing future areas and problems are presented for the community to investigate.

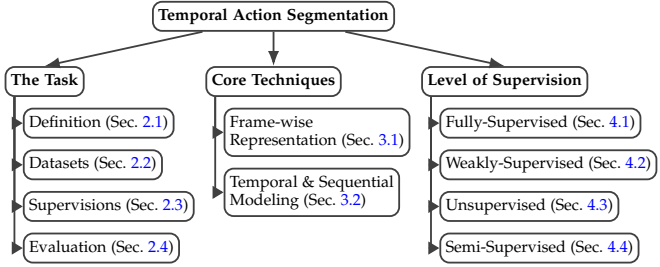
\begin{figure}[!tb]
\centering
{\scriptsize
\resizebox{0.48\textwidth}{!}{ 
\begin{forest}
  dir tree switch=at 1,
  for tree={
    rect,
    align=left,
    edge+={thick, draw=darkgray},
    where level=0{
      colour me out=white,
    }{
      if level=1{
        colour me out=white,
      }{
        colour me out=white, 
        edge+={-Triangle},
      },
    },
  }
  [\textbf{Temporal Action Segmentation }
    [\textbf{The Task}
      [Definition (Sec. \ref{sec:definition})]
      [Datasets (Sec. \ref{sec:datasets})]
      [Supervisions (Sec. \ref{subsec:sup})]
      [Evaluation (Sec. \ref{subsec:evaluation})]
    ]
    [\textbf{Core Techniques}
      [Frame-wise \\Representation (Sec. \ref{sec:encoding})]
      [Temporal \&  Sequential\\ Modeling (Sec. \ref{sec:temporal})]
    ]
    [\textbf{Level of Supervision}
      [Fully-Supervised (Sec. \ref{sec:full})]
      [Weakly-Supervised (Sec. \ref{sec:weak})]
      [Unsupervised (Sec. \ref{sec:unsup})]
      [Semi-Supervised (Sec. \ref{sec:semi})]
    ]
  ]
\end{forest}}}
\caption{The taxonomy of existing TAS research. }\label{fig:structure}
\end{figure}

\textbf{Survey Structure.} 
\cref{fig:structure} outlines a taxonomy of the temporal action segmentation task and the structure of this survey. \cref{sec:task} provides a formal task description and compares it with other related tasks. \cref{sec:datasets,sec:supeval} compares the benchmarks, forms of supervision, evaluation metrics and settings, respectively. \cref{sec:encoding} delves into how frames are embedded and embellished, summarizing the widespread usage of handcrafted models or deep learning backbones for feature extraction. \cref{sec:temporal} outlines the temporal and sequential modeling techniques. \cref{sec:full,sec:weak,sec:unsup,sec:semi} provide a comprehensively curated list of approaches grouped according to the type of supervision. Finally, \cref{sec:conclusion} concludes the survey by discussing challenges and future research directions.

\forestset{%
  colour me out/.style={outer color=#1!75, inner color=#1!50, draw=darkgray, thick, rounded corners},
  rect/.append style={rectangle, rounded corners=1pt},
  dir tree switch/.style args={at #1}{%
    for tree={
      edge=-Latex,
      fit=rectangle,
    },
    where level=#1{
      for tree={
        folder,
        grow'=0,
      },
      delay={child anchor=north},
    }{},
    before typesetting nodes={
      for tree={
        content/.wrap value={\strut ##1},
      },
      if={isodd(n_children("!r"))}{
        for nodewalk/.wrap pgfmath arg={{fake=r,n=##1}{calign with current edge}}{int((n_children("!r")+1)/2)},
      }{},
    },
  },
}

\section{A Primer on Segmentation}\label{sec:task}
This section delves into technical foundations, explores popular datasets used for training and evaluation, examines different levels of supervision, and highlights the evaluation metrics used to assess the performance. 

\subsection{Definition}\label{sec:definition} 
Temporal action segmentation %
partitions a temporally untrimmed video in time and assigns each segment with a pre-defined action label~\cite{richard2017weakly}. Formally, given a video $\bm{x} = (x_1, x_2, ..., x_T)$ of length $T$ with $N$ actions, segmentation methods produce the following output:
\begin{equation}\label{eq:taskseg}
 s_{1:N} = (s_1, s_2, ..., s_N),
\end{equation}
where $s_n = (c_n, \ell_n)$ represents a video segment of length $\ell_n$ with the label $c_n \in \mathcal{C}$, of $\mathcal{C}$ pre-defined categories, and any $\{s_{n},s_{n+1}\}$ segments are consecutive in time. The task can also be regarded as a 1D analogue of semantic (image) segmentation, and be formulated as a frame-wise action classification, \ie, 
\begin{equation}\label{eq:taskframe}
 y_{1:T} = (y_1, y_2, ..., y_T) 
\end{equation}
where $y_t$ is the action label of frame $t$. The segment formulation in \cref{eq:taskseg} is commonly used in weakly supervised works that predict the most probable sequence of actions~\cite{richard2018neuralnetwork,richard2018action}, while the frame-wise formulation of \cref{eq:taskframe} is commonly used with fully supervised methods~\cite{farha2019ms} where dense labels are available. The two formulations are equivalent and one can be reconstructed from the other.

\subsubsection{Related Tasks}

\begin{table}[!tb]
\centering
\caption{
TAS has its unique position in the task landscape, differentiating the tasks based on whether they involve {Temporal Relation}s between action instances in videos, {Boundary Localization} of actions, {Semantic Segment} understanding, and the {Data Domain}.
}\label{tab:comp}
\resizebox{\linewidth}{!}{ 
\begin{tabular}{|l|c|c|c|c|}
\hline
\multirow{2}{*}{\textbf{Task}}   & \textbf{Temporal} &  \textbf{Boundary}   & \textbf{Semantic} & \textbf{Data} \\ 
& \textbf{Relation} & \textbf{Localization} & \textbf{Segment}& \textbf{Domain}\\\hline\hline
Temporal Action Segmentation & \cmark & \cmark &\cmark & video\\
Temporal Action Detection/Localization  & \xmark  & \cmark  & \cmark  &   video\\
Sequence Segmentation & \cmark  &\cmark & \cmark &  audio, motion \\
Key-Frame Detection & \cmark  & \xmark  & \cmark  &  video, text \\
Complex Activity Classification &\cmark  & \xmark  & \xmark & video \\ 
Generic Event Boundary Detection & \xmark  & \cmark  & \xmark  & video\\ 
\hline
\end{tabular}}
\end{table}

Several video understanding tasks are closely related to temporal action segmentation. They can be distinguished based on their data domain, identification of segment semantics, and the reasoning of temporal dynamics between segments. The related tasks are described below.

\textbf{Temporal Action Detection/Localization} (TAD/L)~\cite{shou2016temporal,lin2019bmn} detects the start and end of action instances and predicts semantic labels simultaneously. TAD/L works with general videos from everyday life, such as THUMOS14~\cite{THUMOS14}, and learns from temporally sparse action annotations. In contrast, TAS aims to produce frame-wise dense action labels. 
 
\textbf{Sequence Segmentation} (SS) is popular in other domains, including motion capture data~\cite{barbivc2004segmenting,zhou2008aligned,zhou2012hierarchical,kruger2016efficient} and audio signals~\cite{venkatesh2021investigating}. Most approaches are developed to segment individual sequences~\cite {barbivc2004segmenting,zhou2008aligned,zhou2012hierarchical} while some~\cite{fox2014joint} focuses on multiple motion capture recordings simultaneously. However, such data is lower-dimensional and exhibits much less variance than video. 

\textbf{Key-Frame Detection} (KFD) identifies single characteristic frames or key-steps~\cite{zhukov2019cross,zhang2016video,elhamifar2020self,naing2020procedure} for actions. Like TAS, KFD requires modeling the temporal relations between actions; however, it does not aim to determine the boundaries of action transitions.
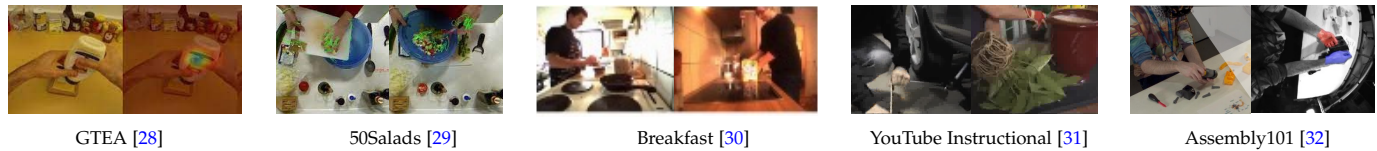
\begin{figure*}[t]
\centering
\begin{overpic}[width=\textwidth]{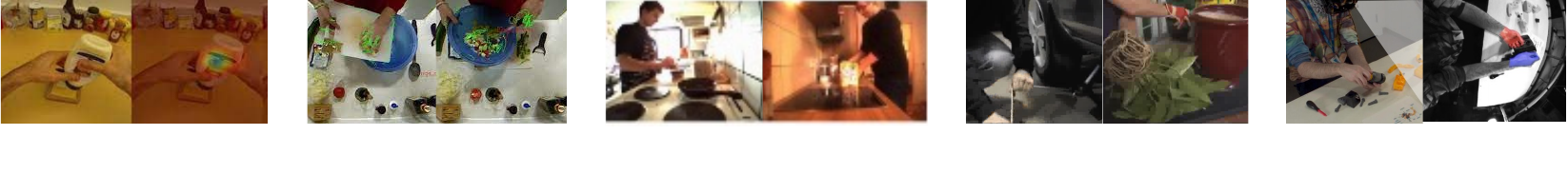} 
    \put(5,1){{GTEA}~\cite{fathi2011learning}}
    \put(25,1){{50Salads}~\cite{stein2013combining}}
    \put(46,1){{Breakfast}~\cite{kuehne2014language}}
    \put(63,1){{YouTube Instructional}~\cite{alayrac2016unsupervised}}
    \put(86,1){{Assembly101}~\cite{sener2022assembly101}}
\end{overpic}
\caption{Images from the core datasets used for TAS. 
The majority of these datasets are from the cooking domain: \eg, GTEA~\cite{fathi2011learning}, 50Salads~\cite{stein2013combining}, and Breakfast~\cite{kuehne2014language}. YouTube Instructional~\cite{alayrac2016unsupervised} comprises tasks from different domains. 
While Assembly101~\cite{sener2022assembly101} focuses on multi-step activities in the assembly  domain.}\label{fig:coredatasets}
\end{figure*}

\begin{table*}[!tb]
\centering
\caption{Comparisons of procedural activity datasets. The first group lists \textbf{Core} datasets for TAS. The second group is commonly used for other \textbf{Related} tasks. We summarize their \textbf{Duration}, \textbf{\# Videos}, \textbf{\# Segments}, \textbf{\# Activity}, \textbf{\# Action}, \textbf{Domain} and \textbf{View}.}
\label{tab:datasetStat} 
\resizebox{\textwidth}{!}{ 
\begin{tabular}{|c|lc|c|cccc|l|l|}
\cline{2-10}
\multicolumn{1}{c|}{}& \textbf{Dataset} & \textbf{Year}  & \textbf{Duration} & \textbf{\# Videos}& \textbf{\# Segments}& \textbf{\# Activity} & \textbf{\# Action} & \textbf{Domain} & \multicolumn{1}{c|}{\textbf{View}} \\
\cline{2-10}\noalign{\vspace{0.65ex}} 
\hline
\multicolumn{1}{|c|}{\multirow{5}{*}{\rotatebox[origin=c]{90}{\textbf{Core}}}} &\cite{fathi2011learning} GTEA & 2011 & 0.4h & 28 & 0.5K &7 & 71 & Cooking   & Egocentric \\ 
&\cite{stein2013combining} 50Salads& 2013 & 5.5h & 50 & 0.9K &1 & 17 & Cooking   & Top-view \\ 
&\cite{kuehne2014language} Breakfast & 2014  & 77h & 1712 & 11K &10 & 48 & Cooking & 3rd Person \\ 
&\cite{alayrac2016unsupervised} YouTube Instructional & 2016 & 7h & 150 & - & 5 & 47 & Mixed  & Mixed  \\ 

&\cite{sener2022assembly101}  Assembly101 &  2022 & 513h & 4321 & 1M &15 & 202  & Assembly  & Egocentric + 3rd Person  \\
\hline\hline 
\multicolumn{1}{|c|}{\multirow{6}{*}{\rotatebox[origin=c]{90}{\textbf{Related}}}}
&\cite{zhou2018towards} YouCookII & 2018 & 176h & 2K & 15K & 89 & - & Cooking & Mixed \\ 
&\cite{zhukov2019cross} CrossTask & 2019  & 376h & 4.7K & 34K & 83 &107 & Mixed  & Mixed \\ 
&\cite{tang2019coin} COIN & 2019 & 476h & 11.8K & 46K &180 &778 & Mixed  & Mixed\\
&\cite{damen2022rescaling} Epic-Kitchens & 2020 & 200h &700&90K&-&4053&Daily&Egocentric\\
&\cite{ben2021ikea} Ikea ASM &  2021& 35h & 371 & 16K &4 & 33 & Furniture  & 3rd Person  \\ 
&\cite{ragusa2021meccano} Meccano &  2021 &  0.3h & 20  & 8.9K  &  1 & 61  & Assembly & Egocentric \\ 
\hline
\end{tabular}} 
\end{table*}
\textbf{Complex Activity Classification} (CAC)~\cite{hussein2019timeception,hussein2020pic} classifies procedural activity videos at the complex activity level. This task is applied to the same data as TAS, but %
does not work at frame-wise resolution. %

\textbf{Generic Event Boundary Detection} (GEBD)~\cite{shou2021generic} %
identify moments in video that humans perceive as class-agnostic event boundaries, indicating changes in action, subject, and environment. Unlike TAS, GEBD does not involve semantic labels or make assumptions about the temporal relations between the detected boundaries.

\cref{tab:comp} compares these tasks to TAS. TAS occupies a unique space in that it requires temporal and sequential reasoning and produces dense frame-wise labels.

\subsection{Datasets}\label{sec:datasets} 
This section discusses the core action segmentation datasets and highlights closely related procedural video datasets currently not used. The core datasets are then analysed based on their size, domain, view, etc., and analyzed for temporal dynamics, action duration distribution, and more.

\subsubsection{Core Datasets}
The core datasets %
records people performing %
procedural activities such as preparing a meal or assembling furniture. The datasets described below are annotated with %
the start, end and class labels of action segments. \cref{fig:coredatasets} shows example images from these datasets. %

\textbf{GTEA}~\cite{fathi2011learning} comprises videos recorded in a single kitchen. The videos are recorded with a camera mounted on a cap worn by four participants. 

\textbf{50Salads}~\cite{stein2013combining} contains videos preparing two types of mixed salads. The videos are captured from a top-down perspective, showcasing the work surface. Participants follow recipe steps that are randomly selected from a statistical recipe model.

\textbf{Breakfast}~\cite{kuehne2014language} targets ``in the wild'' recordings in the kitchens. The dataset features 52 participants performing ten breakfast-related activities. The videos are recorded using 3 to 5 cameras, all capturing the scenes from a third-person perspective.

\textbf{YouTube Instructional}~\cite{alayrac2016unsupervised} is a curated collection comprising five instructional activities, with 30 videos available for each activity. It is primarily used for unsupervised TAS.

\textbf{Assembly101}~\cite{sener2022assembly101} is a recorded dataset in which 53 participants are tasked with disassembling and reassembling take-apart toys without any provided instructions. The dataset includes fine-grained annotations of hand-object interactions and coarse action labels.

\subsubsection{Related Datasets}
There are several other long-range procedural activity datasets; however, there are several challenges or limitations to using them for temporal action segmentation, primarily due to incompatible annotations. For example, the fine-grained action labels which define movements, \eg, `pick up something', hinders the action dynamics modeling for the high-level tasks.

\textbf{Epic-Kitchens}~\cite{Damen2018EPICKITCHENS,damen2022rescaling} is a large-scale egocentric dataset containing 100 hours of recordings. Although it contains long-range videos, the lack of frame-wise dense action labels limits its suitability for temporal segmentation.

\textbf{Ikea ASM}~\cite{ben2021ikea} records people assembling four IKEA furniture and is annotated with fine-grained action labels.

\textbf{Meccano}~\cite{ragusa2021meccano} records 20 people assembling a toy motorbike featuring only fine-grained actions.

\textbf{YouCookII}~\cite{zhou2018towards} is collected from YouTube featuring cooking videos which are annotated with temporal boundaries of recipe steps and textual descriptions only.

\textbf{CrossTask}~\cite{zhukov2019cross} is a YouTube collection of 18 activities with temporal annotations and 65 related tasks without any temporal annotations. It is primarily used to evaluate weakly supervised segmentation algorithms~\cite{lu2021weakly,lu2022set}.

\textbf{COIN}~\cite{tang2019coin} is collected from YouTube and comprises 180 activities spanning twelve domains such as sports, nursing, vehicles, \etc. A video from COIN has on average four segments, which diminishes the inherent potential for modeling sequence dynamics.

\subsubsection{Dataset Comparison \& Discussion} 
\cref{tab:datasetStat} compares procedural video datasets and categorizes the datasets based on their source, scale, number of actions, and viewpoint.

Datasets are typically recorded from static, third-person, or moving, egocentric views. The static background from third-person view is beneficial for recognizing different actions in the same video while the background change poses challenges for discriminating actions across videos. %
Egocentric views excel in capturing objects and tools, facilitating hand-object interaction recognition which is what we are primarily interested for procedural activity understanding. 
However, the camera motion associated with egocentric views introduces additional challenges. The Epic-Kitchens dataset~\cite{damen2022rescaling} is a notable example of a large-scale egocentric vision dataset of untrimmed activities in the kitchen. There are also egocentric datasets for cooking activities on a smaller scale~\cite{fathi2011learning,li2018eye}. %
The Breakfast dataset~\cite{kuehne2014language} primarily consists of recordings from multiple third-person viewpoints. Only Assembly101~\cite{sener2022assembly101} provides synchronous egocentric and third-person views among these datasets.
 
Curating videos from online platforms like YouTube is convenient for creating large-scale and diverse datasets~\cite{malmaud2015s,zhou2018towards,zhukov2019cross,tang2019coin,sener2019zero,miech2019howto100m}. These datasets prove valuable for training offline retrieval systems~\cite{lin2022egocentric} and representation learning~\cite{miech2020end}. However, their applicability may be limited for real-time tasks such as action anticipation or early detection. Specifically, online videos are often produced content; there may be domain gaps that arise from editing processes like fast-forwarding, annotated frames, or changing viewpoints. 

Yet, the diversity of existing recorded procedural activity datasets is rather limited. With only a few small-scale exceptions~\cite{ben2021ikea,ragusa2021meccano}, most recorded datasets ~\cite{stein2013combining,kuehne2014language} focus exclusively on cooking and kitchen activities. Presently, Assembly101~\cite{sener2022assembly101} stands as the only large-scale dataset that extends beyond the cooking domain.

\subsubsection{Background Frames} 
Some videos feature task-irrelevant segments. For example, the actor may talk or give recommendations without performing actions of interest, introduce tools, or demonstrate alternative ways to complete an action. Such \emph{background frames} occur at arbitrary locations with varying lengths and are common in datasets collected from YouTube, such as YouTube Instructional~\cite{alayrac2016unsupervised}, YouCookII~\cite{zhou2018towards}, and CrossTask~\cite{zhukov2019cross}. In most existing works, the background class is treated equally as other action classes and used during training and inference.

\subsubsection{Temporal Dynamics}
\begin{table}[t]
\centering
\caption{Temporal dynamics of three TAS datasets. A higher value in \textbf{Repetition} score indicates relatively more action repetition. A lower value in \textbf{Order Variation} score indicates looser action ordering constraints. }\label{tab:tempdynamics}
\begin{tabular}{|l|c|c|}\hline
\multicolumn{1}{|l|}{\textbf{Dataset}} &\textbf{Repetition $r$} $\uparrow$ &\textbf{Order Variation $v$} $\downarrow$  \\\hline\hline
\cite{stein2013combining} 50Salads &0.08 & 0.02\\
\cite{kuehne2014language} Breakfast & 0.11 & 0.15\\ 
\cite{sener2022assembly101} Assembly101 & 0.18 & 0.05 \\\hline
\end{tabular}
\end{table}

A defining characteristic of the action segmentation task is the temporal dynamics between actions. To qualitatively assess the temporal dynamics, we define a repetition score and an order variation score. The extent of actions repeating in a sequence is quantified by the \textbf{repetition score}, $r$, defined as
\begin{equation}\label{eq:rep}
r = 1 - u / g,
\end{equation}
where $u$ is the number of unique actions, $g$ is the total number of actions in a sequence. The score $r$ ranges between 0 and 1, where 0 signifies no repetitions, and higher scores reflect a higher degree of repetition within the sequence.

The \textbf{order variation score}, $v$, is defined as the normalized average edit distance, $e(X,Y)$, between a pair of sequences, $(X,Y)$. This score is further normalized with respect to the maximum sequence length.
\begin{equation}\label{eq:ov}
v = 1-e(X,Y)/\text{max}(|X|, |Y|),
\end{equation}
The score $v$ also has a range of $[0,1]$, where $1$ indicates no deviations in ordering between action pairs. Conversely, a lower score indicates a higher amount of ordering variations. 
Assembly101~\cite{sener2022assembly101} positions itself as a challenging benchmark for modeling the sequence dynamics between actions. As shown in~\cref{tab:tempdynamics}, Assembly101's order variation score falls between that of Breakfast~\cite{kuehne2014language} and 50Salads~\cite{stein2013combining} and includes relatively more repeated steps than the two datasets respectively (see Table~\ref{tab:tempdynamics}).

\definecolor{bars}{RGB}{130,179,102}
\begin{figure}
\centering
\includegraphics{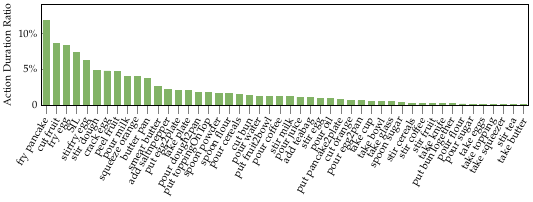}
\caption{Action duration distribution on Breakfast~\cite{kuehne2014language}, sorted by descending order. `SIL' indicates the `background' where no action of interest occurs. The head class `fry pancake' is $639 \times$ more frequent than the tail class `take butter'.}\label{fig:breakfast}
\end{figure}

\subsubsection{Long-Tailed Action Distributions}\label{subsec:frame_freq}
Procedural videos feature a wide range of actions; the actions occur naturally with differing frequencies and durations. For example, `pour coffee' is more common than the optional step of `pour sugar' in making coffee and `fry egg' requires considerably more time than `crack egg' in making fried egg. The long-tailed distribution of action occurrence and duration is an overlooked aspect of per-frame classification formulation~(\cref{eq:taskframe}) of temporal action segmentation. To quantify the action duration discrepancies, we calculate the proportion of each action label within the entire dataset, \ie, $n_c/\sum_{i=1}^{\mathcal{C}}n_i$, where $n_c$ is the number of frames with label $c$. Figure \ref{fig:breakfast} depicts the imbalanced distribution of action duration in Breakfast~\cite{kuehne2014language}.
 
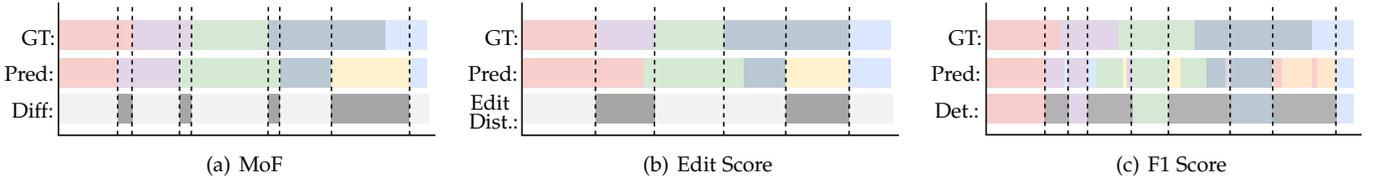
\begin{figure*}[tb]
    \centering
    \hspace{1.5em}
    \subfigure[MoF]{\label{subfig:acc}
    \begin{overpic}[width=0.28\textwidth]{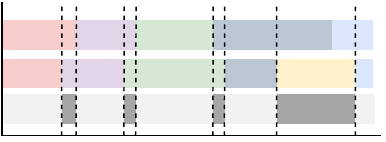}
        \put(-9,25){\footnotesize GT:}
        \put(-13.5,15.8){\footnotesize Pred:}
        \put(-11.6,6){\footnotesize Diff:}
    \end{overpic}}
    \hfill
    \subfigure[Edit Score]{
    \begin{overpic}[width=0.28\textwidth]{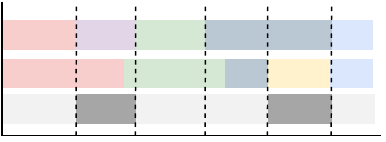}
        \put(-9,25){\footnotesize GT:}
        \put(-13.4,15.8){\footnotesize Pred:}
        \put(-12.8,8.25){\footnotesize Edit}
        \put(-13.2,3.25){\footnotesize Dist.:}
    \end{overpic}}
    \hfill
    \subfigure[F1 Score]{
    \begin{overpic}[width=0.28\textwidth]{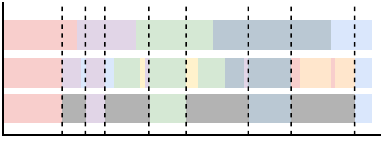}
        \put(-9,25){\footnotesize GT:}
        \put(-13.5,15.8){\footnotesize Pred:}
        \put(-12.6,6){\footnotesize Det.:}
    \end{overpic}}
    \label{fig:enter-label}
    \caption{Evaluation metric visualization. (a) \textbf{MoF} estimates how accurate frame wise predictions are. (b) \textbf{Edit score} tolerates small boundary shifts, as long as the sequence order is correct. (c) \textbf{F1 score} is the harmonic mean of the precision and recall of the detected action segments. The formulas can be found  in~\cref{subsec:evaluation}. %
    }
    \label{fig:evalviz}
\end{figure*}

\begin{table}[t]
\centering
\caption{Imbalance Ratio (IR) on four TAS datasets.}\label{tab:long}
\begin{tabular}{|l|c|c|c|c|}\hline
\multicolumn{1}{|l|}{\textbf{Dataset}} &\textbf{GTEA} &  \textbf{50Salads}& \textbf{Breakfast} & \textbf{Assembly101} \\\hline\hline
\multicolumn{1}{|l|}{\textbf{IR}} &24& 6 & 639& 2604\\\hline
\end{tabular}
\end{table}

The imbalance ratio (IR) is a widely used metric to quantify the skew in datasets~\cite{liu2019large,kang2019decoupling}. IR is defined as the ratio between the number of frames in the head and the tail classes ($n_1/n_K$) sorted by the decreasing order of cardinality (\ie, if $i_1\!>\!i_2$, then $n_{i_1}\! \geq\! n_{i_2}$). \cref{tab:long} shows that 50Salads~\cite{stein2013combining} has the smallest IR value of 6, indicating the most frequent action class has only 6 times the number of frames compared to the least frequent class. On the other hand, Assembly101~\cite{sener2022assembly101} is highly imbalanced in action duration distribution with an IR of 2604. Such a long-tailed nature of the datasets poses extra challenges to the TAS task.

\subsection{Supervision}\label{subsec:sup}

TAS has been investigated under different forms of supervision. A fully-supervised setting provides dense action labels for every frame in training video sequences~\cite{farha2019ms,lea2017temporal}. Dense labels are the most time-consuming to collect per video sequence as it requires the annotator to view the entire video sequence~\cite{moltisanti2019action}. 
A semi-supervised~\cite{singhania2022iterative,ding2022leveraging} setting reduces the annotation effort proportionally by annotating a subset of the videos densely while treating the remaining videos as unlabeled samples.

Weak labels require less annotation effort than dense video labels. %
Time-stamp annotations~\cite{li2021action,rahaman2022generalized} are sparsely labelled single frames interspersed through a sequence and can be viewed as an ordered list of actions associated with exemplar frames. A weaker form is the action list or action transcript~\cite{richard2018neuralnetwork,li2019weakly}, which does away with the exemplar frames. An even weaker form is the action set, which does away with the ordering of actions and provides only the set of all action labels present in the video~\cite{richard2018action,fayyaz2020sct,li2020set,lu2022set}. More recently,~\cite{ding2021temporal} showed that one can do away with action-level annotations and simply use the video-level complex activity label for supervision. 

The unsupervised setting in TAS works~\cite{sener2018unsupervised,sarfraz2021temporally,du2022fast} considers collections of videos that perform the \emph{same} activity. In this regard, it is not label-free, as it requires the activity label to form the video collections. The unsupervised setting is therefore comparable with the weak activity label supervision of~\cite{ding2021temporal} regarding label information. However, the two settings differ in how the collections of videos are processed during training. Formally, unsupervised works work with one group of the same activity videos at a time, while activity label supervision works with videos from all activities simultaneously.

\subsection{Evaluation}\label{subsec:evaluation}

\subsubsection{Evaluation Measures}
Three commonly adopted evaluation metrics in TAS are Mean over Frames (MoF), Edit Score, and F1 scores. The first is a frame-based measure, while the latter two are segment-based measures. All three metrics are used in full, weak, and semi-supervised settings. For unsupervised settings, only F1 and MoF are reported in the literature. %
Note that the evaluation of the unsupervised works is conditioned on the association between clusters and semantic labels. The community has adopted the Hungarian matching algorithm for this purpose (see details in~\cref{subsec:evalunsup}).

\textbf{Frame-Based Measures.} 
Mean over Frames (MoF) is a frame-wise accuracy %
and is defined as the fraction of the model's correct frame predictions:
\begin{equation}
\text{MoF} = \frac{\text{\# of\ correct\ frames}}{\text{\# of\ all\ frames}}\label{eq:acc}.
\end{equation}

The MoF metric can be problematic under dataset imbalance, \ie if frequent and long action classes dominate. The long-tailed nature of current datasets (see~\cref{subsec:frame_freq}) %
implies that models with similar MoF scores may have large qualitative differences, suggesting that class-balanced metrics may be more appropriate though this is currently not adopted in the literature. %
Futhermore, MoF, as a per-frame calculation, does not capture segment quality; the score can be high even when the segments are fragmented. Dividing an action into many discontinuous sub-segments is referred to as over-segmentation. %
Segment-based measures like Edit-score~\cite{lea2016segmental} and F1-score~\cite{lea2017temporal} are instead more suitable measures for evaluating phenomenon like over-segmentation.

\textbf{Segment-Based Measures.} 
The Edit Score~\cite{lea2016segmental} quantifies the similarity of two sequences. It is based on the Levenshtein or edit distance and tallies the minimum number of insertions, deletions, and replacement operations required to convert one segment sequence into another. Consider $X$ and $Y$ as the ordered list of predicted and ground truth action segments; the accumulated distance value $e$ is defined as:

\begin{equation}
 \!\!\!\!\!\!e[i,j] =
 \begin{cases}
\max(i,j), & \min(i,j)=0\\
 \begin{aligned}
 \min(&e[i\!-\!1,j]\!+\!1,e[i,j\!-\!1]\!+\!1, \\&e[i\!-\!1,j\!-\!1]\!+\!\mathbbm{1}(X_i\!\neq \!Y_j))
 \end{aligned}, & \text{otherwise.}
\end{cases}\label{eq:e}
\end{equation}
where $i \in |X|,j \in |Y|$ are indices for $X$ and $Y$, respectively, and $\mathbbm{1}(\cdot)$ is the indicator function. \cref{eq:e} can be effectively solved by dynamic programming. The final edit distance value is then normalized by the maximum length of the two sequences to compute the Edit score:
\begin{equation}
 \text{Edit Score} = \left( 1 - \frac{e[|X|,|Y|]}{\text{max}(|X|,|Y|)} \right) \cdot 100.
\end{equation}
As a metric, the Edit score can assess how well a model predicts the sequence of actions without requiring exact frame-wise correspondence to the ground truth. 

The F1 score or F1@$\tau$~\cite{lea2017temporal} compares the Intersection over Union (IoU) of each segment with respect to the corresponding ground truth based on some threshold $\tau/100$. A segment is considered a true positive if its IoU with respect to the ground truth exceeds the threshold. %
If there is more than one correct segment within the span of a single ground truth action, then only one segment is considered a true positive and the others are marked as false positives. Based on the true and false positives as well as false negatives (missed segments), one can compute the precision and recall and blend the two into the harmonic mean to get 
\begin{equation}
\text{F1} = 2 \cdot \frac{\text{precision}*\text{recall}}{\text{precision}+\text{recall}}\label{eq:f1}.
\end{equation}
Commonly used $\tau$ values are $\{10,25,50\}$. ~\cref{fig:evalviz} visualizes these three evaluation metrics.

\subsubsection{Hungarian Matching for Unsupervised Evaluation}\label{subsec:evalunsup}
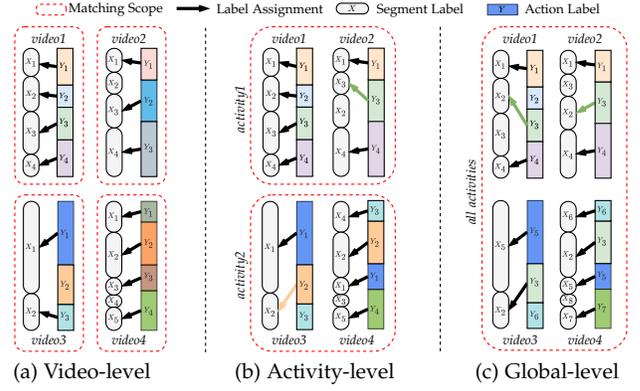
\begin{figure}[tb]
    \centering
    \begin{overpic}[width=0.45\textwidth]{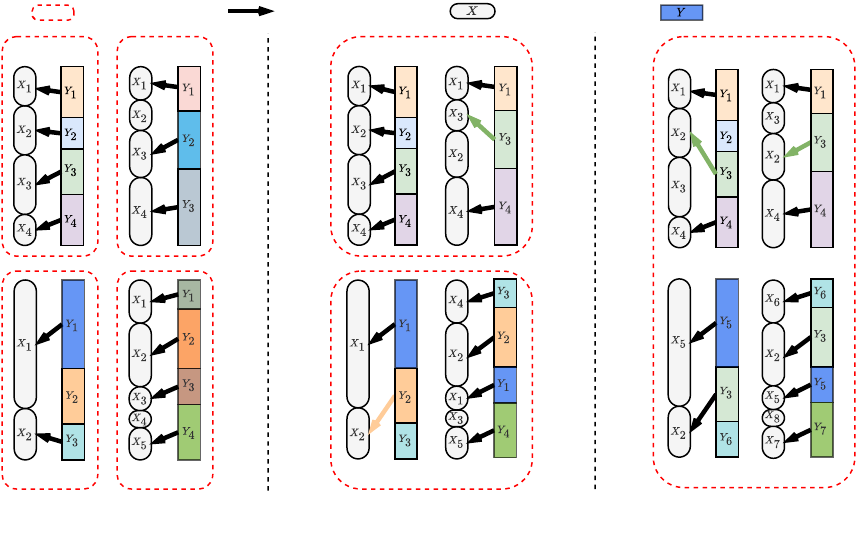}
    \put(9,60.5){\tiny Matching Scope}
    \put(33,60.3){\tiny Label Assignment}
    \put(59,60.3){\tiny Segment Label}
    \put(83,60.3){\tiny Action Label}
    \put(3,55.5){\tiny \textit{video1}}
    \put(16,55.5){\tiny \textit{video2}}
    \put(3,6.5){\tiny \textit{video3}}
    \put(16,6.5){\tiny \textit{video4}}
    \put(42,55.5){\tiny \textit{video1}}
    \put(53,55.5){\tiny \textit{video2}}
    \put(42,6.5){\tiny \textit{video3}}
    \put(53,6.5){\tiny \textit{video4}}
    \put(79.5,55.5){\tiny \textit{video1}}
    \put(90,55.5){\tiny \textit{video2}}
    \put(79.5,6.5){\tiny \textit{video3}}
    \put(90,6.5){\tiny \textit{video4}}
    \put(36,41){\rotatebox{90}{\tiny \textit{activity1}}}
    \put(36,14){\rotatebox{90}{\tiny \textit{activity2}}}
    \put(73.5,26){\rotatebox{90}{\tiny \textit{all activities}}}
    \put(0,1){\footnotesize (a) Video-level}
    \put(36,1){\footnotesize (b) Activity-level}
    \put(75,1){\footnotesize (c) Global-level}    
    \end{overpic}
    \caption{Four videos from two complex activities with varying Hungarian matching levels.  Colored rectangles denote ground truth actions, while rounded rectangles denote video segments. Hungarian matching scopes are red dashed rectangles. Black dashed arrows indicate matched segments, while the coloured arrows highlight changed matching across levels. With scope changing from the video~(a) to activity~(b), change in GT $Y$ across different videos results in the change of label association for \textit{video2}~(green) and \textit{video3}~(orange). A similar change of assignments happens when the matching is done on the global level~(c). Unmatched segments ($X_2$ in \textit{video2} at video-level matching~(a)) are considered as background.}\label{fig:hungarian}
\end{figure}

The evaluation metrics in~\cref{subsec:evaluation} are not directly applicable to the unsupervised scenario without some correlation between the estimated agnostic segments and ground truth actions. 
The Hungarian matching algorithm~\cite{kuhn1955hungarian} is a combinatorial algorithm used to find maximum-weight matching in bipartite graphs, and it is widely adopted for evaluating unsupervised clustering tasks~\cite{li2006relationships,chang2019deep}. 
In unsupervised TAS, Hungarian matching links the given frames $X$ of $N$ clusters to the action label corpus $Y$ of $M$ classes with the best matching $\widehat{\assignment} \subset \{0,1\}^{N\times M}$:
\begin{equation}
 \begin{aligned}
 \widehat{\assignment} = \argmax_{\assignment} \sum_{n,m} \mathcal{A}_{n,m} \cdot I(X_n,Y_m),\\
 \text{s. t.} \quad |\assignment| = \min(N,M)
 \end{aligned}
\end{equation}
where $X_n$ denotes frames belonging to cluster $n$, and $Y_m$ denotes frames with the action label $m$. $\assignment_{n,m}$ is the indicator function for assigned pair $(n, m)$, $I(X_n,Y_m)$ is the number of frames with ground-truth class label $m$ that appear in cluster $n$. 
When two sets have equivalent classes ($N\!\!=\!\!M$), the Hungarian matching constructs a bijection. Otherwise, it produces a one-sided perfect matching of size $min(N,M)$. The remaining mismatched clusters are treated as background automatically. The evaluations are then based on the corresponding results. Depending on the bipartite set's scope, Hungarian matching can be applied at three different levels, as illustrated in~\cref{fig:hungarian}.

\textbf{Video-level} matching~\cite{aakur2019perceptual,sarfraz2021temporally} matches the cluster with respect to the ground truth actions of a \emph{single} video. This matching evaluates the ability of a model to segment a video sequence into distinct actions and produces the highest evaluation scores because of the limited scope. Within each matching scope, as shown in \cref{fig:hungarian}(a), the matching is agnostic to associations across videos.

\textbf{Activity-level} matching associates clusters to labels within each complex activity. 
Most unsupervised works~\cite{sener2018unsupervised,kukleva2019unsupervised,vidalmata2021joint} follow this scope of matching, \ie, process videos from the same activity. As shown in \cref{fig:hungarian}(b), the activity level of grouping leads to the assignment changes denoted by the coloured arrows. 

Lastly, \textbf{global-level} matching is performed on the entire dataset. Introduced in~\cite{ding2021temporal}, it is the most challenging setting, as both intra- and inter-activity matching must be considered. It is noteworthy that~\cite{kukleva2019unsupervised} reports different `global' matching results across complex activities, as their setting does not consider actions shared across complex activities. 

The various scopes of Hungarian matching correlate to distinct learning objectives of a TAS model; the greater the scope, the more general the task. Video-level matching sets the requirement of differentiating actions from one another within a video, \ie, \emph{intra-video action discrimination}. For activity-level matching, 
a model must discriminate between actions within a video and form \emph{intra-activity action association}. 
In the global-level matching, a model must include \emph{inter-activity associations} to construct feasible action correspondences across complex activities. 
Note that a model learned at a broader scope is \textbf{downwards compatible} and can be adjusted for evaluation at a finer scope, \eg, from global to activity level, but not vice versa. Despite the practical feasibility of doing so, the results are not directly comparable due to the models' disparate learning requirements. %

\section{Core Techniques} 
TAS benefits from two core techniques: \textbf{frame-wise representations}, which extracts informative features at a frame-level to capture spatial appearance and motion information, and \textbf{temporal and sequential modeling}, which incorporates temporal dependencies and sequential context for improved TAS accuracy.
\subsection{Frame-Wise Representations}\label{sec:encoding}
The standard practice in TAS is to use pre-computed frame-wise features, without end-to-end learning, %
as inputs. This convention is due to the heavy computational demands of learning video features. Using pre-computed features has a key advantage in that it allows a dedicated comparison of the proposed architectures without the confounding influences of improved frame-wise feature representations. %

\textbf{Fisher Vector Encoded IDT.}~\label{sec:idt} The original and Improved Dense Trajectories (IDT)~\cite{wang2011action,wang2013action} were commonly used hand-crafted features for action recognition and video understanding before the rise of deep learning. The original dense trajectories features~\cite{wang2011action} are spatiotemporal features computed along tracks of interest points formed via optical flow. IDT~\cite{wang2013action} corrects the trajectories for camera motions. To apply IDT to action recognition,~\cite{wang2013action} further encode the raw trajectories by using Fisher Vectors (FV)~\cite{perronnin2010improving} to capture the trajectories' first and second-order statistics.

\textbf{Inflated 3D ConvNet (I3D).}~\label{sec:i3d}
I3D~\cite{carreira2017quo} is a state-of-the-art architecture to extract generic features for video understanding. It uses as a backbone the pre-trained Inception-V1~\cite{ioffe2015batch} with 2D ConvNet inflation. In practice, it inflates all $N\!\times\!N$ spatial kernels to $N\!\times\!N\times\!N$ by replicating the original kernels $N$ times and rescaling them with a temporal factor of $1/N$. The model is pre-trained on the Kinetics dataset~\cite{kay2017kinetics} for action recognition. Architecture-wise, the I3D model has two data streams \ie, RGB and optical flow. The optical flow of the input video is computed by the TV-L1 algorithm~\cite{zach2007duality}. Then, a $21\!\times\!224\!\times\!224$ spatiotemporal volume of RGB and flow frames are each fed into their respective branches to extract 1024D features. The two are then concatenated to compose the final 2048D representation. %

\begin{figure}[tb]
    \centering
    \begin{overpic}[width=\linewidth]{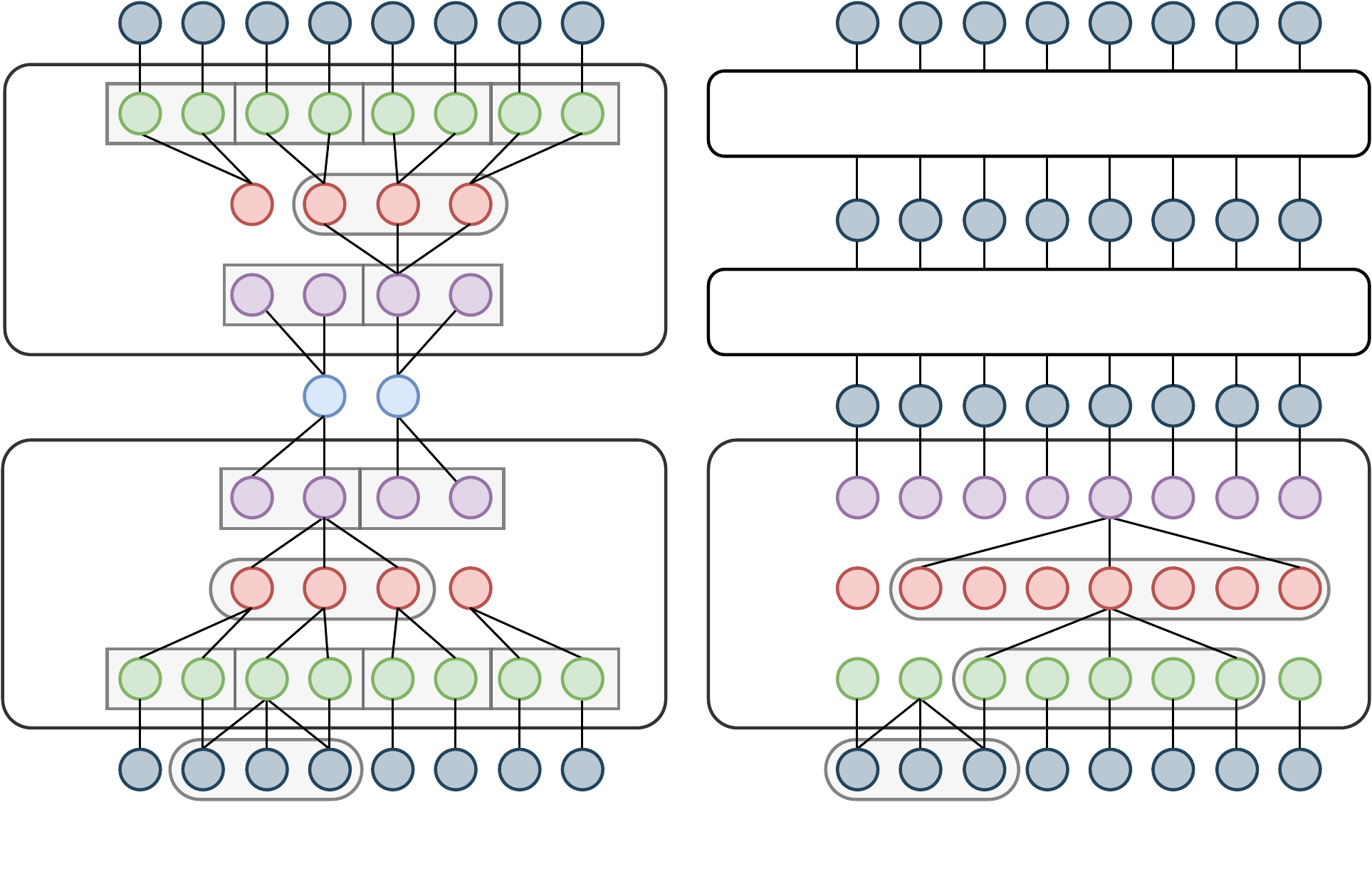}
    \put(1,7){\tiny input}
    \put(1,12){\tiny conv}
    \put(1,18){\tiny pool}
    \put(1,24){\tiny conv}
    \put(1,30){\tiny pool}
    \put(1,40){\tiny upsample}
    \put(1,46){\tiny conv}
    \put(1,52){\tiny upsample}
    \put(0,61){\tiny output}
    \put(37.5,28){{Encoder}}
    \put(37.5,40){{Decoder}}
    \put(72,40){{Stage 2}}
    \put(72,55){{Stage 3}}
    \put(52.5, 7){\tiny input}
    \put(52.5,12){\tiny rate=0}
    \put(52.5,18){\tiny rate=1}
    \put(52.5,24){\tiny rate=2}
    \put(51,34){\tiny output1}
    \put(51,47){\tiny output2}
    \put(51,61){\tiny output3}
    \put(8,0){ (a) Encoder-decoder TCN}
    \put(61,0){ (b) Multi-stage TCN}
    \end{overpic}
    \caption{Two exemplary types of Temporal Convolutional Networks for TAS. (a) Encoder-decoder TCNs progressively enlarge the temporal receptive field via pooling. (b) Multi-stage TCNs maintain a fixed temporal resolution with progressively larger dilated convolutions.}\label{fig:tcn}
\end{figure}
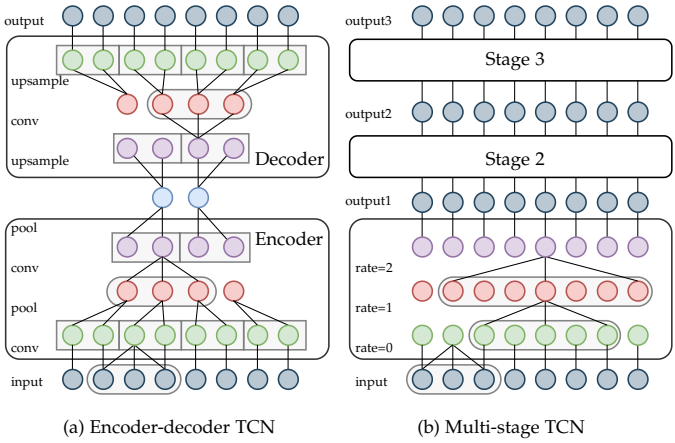

\subsection{Temporal and Sequential Modeling}\label{sec:temporal} 
Segmenting actions according to frame-wise features outlined in \cref{sec:encoding} typically requires some additional handling on the overall sequence. %
In accordance with the hierarchical structure of these videos, the reasoning of the temporal dynamics can be categorized into frame-level and segment-level. We denote the frame-level model as temporal modeling and the segment-level model as sequential modeling. 

\subsubsection{Temporal Modeling}
Temporal modeling on a frame-wise basis expands the temporal receptive field of the network and aggregates the dynamics in the feature representations. This level of modeling allows for information exchange across the frame-wise feature representations. 
Efforts dedicated to the temporal modeling include Recurrent Neural Networks, Temporal Convolutional Networks, and Transformers.

\textbf{Recurrent Neural Networks (RNNs).} 
RNNs %
capture the temporal relations recurrently with a set of parameters shared over time. 
Among the RNN variants, uni-directional~\cite{richard2017weakly} and bi-directional~\cite{huang2020improving} Gated Recurrent Units (GRUs)~\cite{cho2014learning} are used. Specifically, the network takes in input features recurrently following their temporal order and predicts action labels. %
The memory capacity of a frame-wise RNN or GRU, however, does not span long enough to capture the sequential relationship between actions. The discussed methods~\cite{richard2017weakly,huang2020improving} are therefore usually combined with sequential modeling techniques introduced in~\cref{subsec:seqmodel}. Another weakness of an RNN is its limited ability to process sequential inputs in parallel due to the recurrent dependencies between frames.

\textbf{Temporal Convolutional Networks (TCNs).}~\label{sec:TCNsection} 
TCNs~\cite{lea2017temporal} use 1D convolutional kernels in time. Two standard paradigms of TCNs, shown in \cref{fig:tcn}, are encoder-decoders and multi-stage TCNs. Encoder-decoder TCNs~\cite{ding2018weakly,lea2017temporal,lei2018temporal,singhania2021coarse} shrink and then expand the temporal resolution, using layer-wise pooling and upsampling in a U-Net fashion~\cite{ronneberger2015u}. In contrast, the multi-stage architecture (MS-TCN) keeps a constant temporal resolution 
and expands the receptive field with progressively larger dilated convolutions~\cite{farha2019ms,li2020ms}. Comparatively, the encoder-decoder architecture can reduce the computation time of long input sequences with temporal pooling. However, pooling may harm the prediction accuracy at action boundaries. The MS-TCN architecture preserves the full temporal resolution, especially the boundary information, but comes at a cost of higher computation.

\begin{figure}
    \centering
    \begin{overpic}[width=\linewidth]{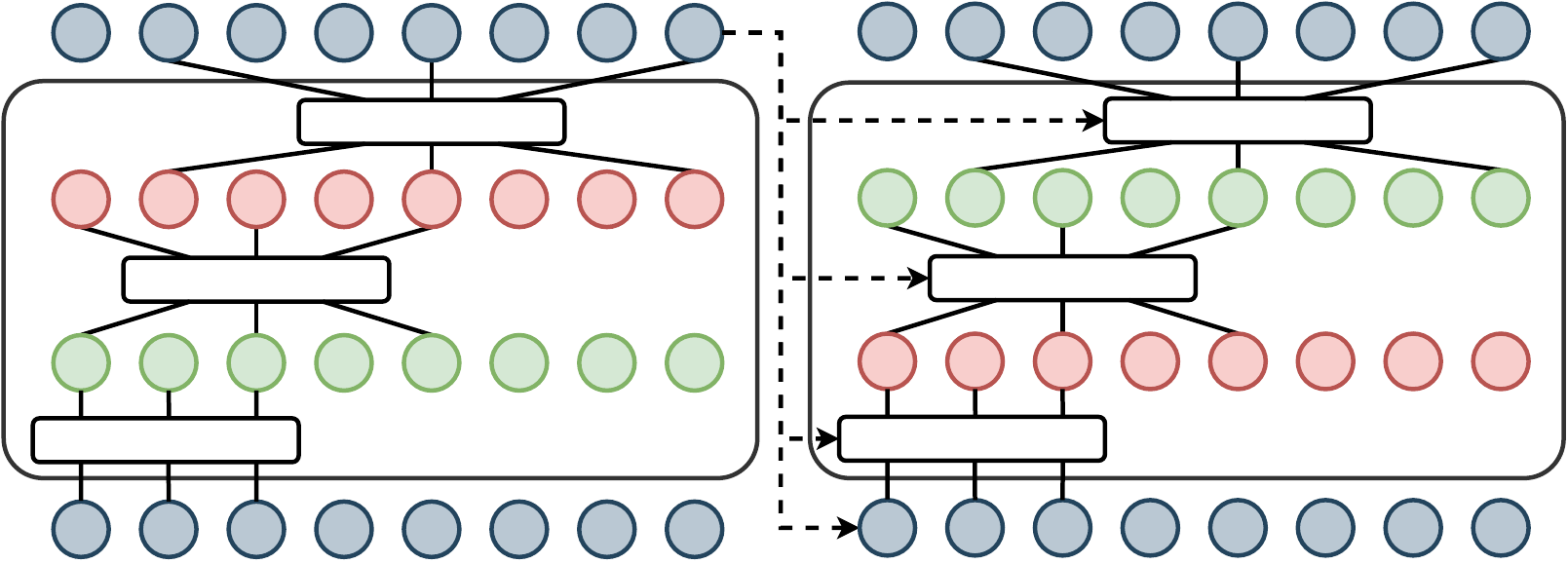}
    \put(6,7.4){\tiny TB w/ SA}
    \put(11.5,17.5){\tiny TB w/ SA}
    \put(23.5,27.6){\tiny TB w/ SA}
    \put(35,7){Encoder}
    \put(57,7.4){\tiny TB w/ CA}
    \put(62.5,17.5){\tiny TB w/ CA}
    \put(74.5,27.6){\tiny TB w/ CA}
    \put(86,7){Decoder}
    \end{overpic}
    \caption{Transformer architecture for TAS. The transformer blocks (TBs) take as input frames with increasing temporal dilation ratios $d$. TBs in encoder use self-attention (SA) across the frames. The decoder takes as input encoder outputs and applies cross-attention (CA) in TBs at each layer.}
    \label{fig:transformer}
\end{figure}
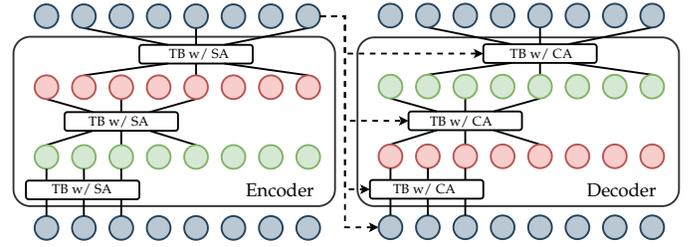

\textbf{Transformer.} 
Transformers~\cite{vaswani2017attention} 
have seen a quick adoption for video tasks, including TAS; we refer the reader to a recent survey~\cite{ulhaq2022vision}. %
The core technique of a transformer is the attention mechanism; Sener~\etal~\cite{sener2020temporal} proposed one of the first attention-based architectures. Named Temporal Aggregates, this model uses a non-local operation~\cite{nonlocalNetVLAD} for %
mutual attention between frames in multiple time spans. %

ASFormer~\cite{yi2021asformer} was one of the first true transformer architectures for TAS. %
It translates the encoder-decoder architecture of the ED-TCN~\cite{lea2017temporal} by replacing the convolutional operations with transformer blocks. %
The encoder uses pre-computed I3D features as input and self-attends to the frame-wise features within the inputs, while the blocks in the decoder adopt cross-attention %
between features and the encoder outputs (see~\cref{fig:transformer}).
Building on ASFormer, Behrmann~\etal~\cite{behrmann2022unified} adjust the decoder to output only the action sequence instead of frame-wise action labels, \ie, mapping frame inputs to action sequence outputs. 

Transformers are gradually being embraced for TAS, but their use is still limited. First, the transformers lack inductive biases and require large video corpora for effective training. Yet existing datasets for TAS are relatively small, making it difficult for large transformers to develop effective representations. Another issue identified by~\cite{zhu2020deformable} is that the self-attention mechanism might not acquire meaningful weights from large spans of inputs.

In summary, 
regardless of the chosen network, the temporal relationships for TAS is modeled implicitly through the network architecture. %

\subsubsection{Sequential Modeling}\label{subsec:seqmodel}
The actions in procedural videos typically follow some %
ordering to %
achieve a specific goal. Such sequential information is more easily captured on a segment level. Various sequence models, such as Hidden Markov Models and Mallows Models, have been investigated~\cite{richard2018neuralnetwork,sener2018unsupervised,kukleva2019unsupervised}.

\textbf{Hidden Markov Models (HMMs).}\label{subsubsec:hmm} 
HMMs are classic probabilistic models for working with sequential data. %
Recall that for a given video $\bm{x}$, the objective is to find %
segments $(\hat{\bm{c}},\hat{\bm{\ell}})$, where $\hat{\bm{c}} = [ \hat{c}_1,\hdots, \hat{c}_n,\hdots, \hat{c}_{\hat{N}} ]$ denotes the predicted ordering of action labels of length $\hat{N}$ and $\hat{c}_n\in\mathcal{C}$, and $\hat{\bm{\ell}} = [\hat{\ell}_1,\hdots, \hat{\ell}_{\hat{N}}]$ are their corresponding temporal extents. The HMM estimates the MAP result $(\hat{N},\hat{\bm{c}},\hat{\bm{\ell}})$, which can be written as:
 \begin{align}\label{eq:hmm}
 & (\hat{N}, \hat{\bm{c}}, \hat{\bm{\ell}})
 =\argmax_{N,\bm{c},\bm{\ell}} p(\bm{c},\bm{\ell}|\bm{x}) \\ 
 & =\argmax_{N,\bm{c},\bm{\ell}} p(\bm{c}) \cdot p(\bm{\ell}|\bm{c}) \cdot p(\bm{x}|\bm{c}) \nonumber\\
 & =\argmax_{N,\bm{c},\bm{\ell}} \underbrace{\left[\prod_{n=1}^{N-1} p(c_{n+1}|c_n)\right]}_{\text{context model}} \underbrace{\left[\prod_{n=1}^N p(\ell_n|c_n)\right]}_{\text{length model}} \underbrace{\left[\prod_{t=1}^T p(x_t|c_n)\right]}_{\text{visual model}}. \nonumber
 \end{align}
The last term $p(\bm{x}|\bm{c})$ in the second line is simplified from $p(\bm{x}|\bm{c},\bm{\ell})$ since it is a frame-wise likelihood and does not depend on the action length $\bm{\ell}$.

The HMM formulation induces a three-component model. The first, $p(\bm{c})$, is a \emph{context model}, providing probabilities for the sequence of actions in the video. The context can be computed from the training data based on the provided action labels. Alternative solutions have also been proposed for the unsupervised setting without labels.
For example, the CTE approach~\cite{kukleva2019unsupervised} assumes that similar actions happen in close temporal vicinity so that the average timestamp $t(k)$ for feature clusters in the temporal embedding space is a good indication of the action order in the sequence:
\begin{equation}
 t(k) = \frac{1}{|\mathbf{F}(k)|} \sum_{f\in \mathbf{F}(k)} t(f),
\end{equation}
where $\mathbf{F}(k)$ is the set of features in cluster $k$ and $t(\cdot)$ indicates the normalized temporal location. The clusters are then ordered as $\bm{\pi}=[k_1,\dots,k_K]$ with respect to their temporal location, such that $0\leq t(k_1)\leq \dots \leq t(k_K) \leq 1$. With this ordering, the transition probability is defined as:
\begin{equation}\label{eq:rigid}
 p(c_{n+1}|c_n) = 
 \begin{cases}
 1, & c_{n+1}= c_n ~\text{or}~k_{c_{n+1}} - k_{c_n}=1, \\
 0, & \text{otherwise.}
 \end{cases}
\end{equation}
\cref{eq:rigid} imposes a hard transition; %
new frames must remain the same action label as the previous frame or transition to the next action label observed in the ordering $\bm{\pi}$. Li~\etal~\cite{li2021action} define a relaxed alternative, taking into consideration the action length $\lambda_c$:
\begin{equation}\label{eq:relax}
 p(c_{n+1}|c_n) \propto
 \begin{cases}
 \frac{\lambda_{c_n}+\lambda_{c_{n+1}}}{\sum_{j=c_n}^{c_{n+1}}\lambda_{c_j}}, & k_{c_{n+1}}>k_{c_n}, \\
 0, & \text{otherwise.}
 \end{cases}
\end{equation}
This formulation allows for skipping actions in the ordering $\bm{\pi}$ and penalizes multiple action skips with a large denominator (sum of skipped action lengths) in \cref{eq:relax}.

The second component $p(\bm{\ell}|\bm{c})$, the \emph{length model}, determines the temporal length for each action class. The common practice~\cite{richard2018neuralnetwork,richard2018action,li2020set,li2021action} is to model the length of each action with a Poisson distribution:
\begin{equation}
 p(\ell|c) = \frac{\lambda_c^\ell}{\ell!}e^{-\lambda_c}.
\end{equation}
The lengths $\lambda_c$ for actions are estimated over all the video sequences by the following:
\begin{equation}\label{eq:len}
 \hat{\lambda} = \argmin_{\lambda} \sum_{x\in X}\left(\sum_{c\in \mathcal{A}_x}\lambda_c-T_x\right)^2, \quad \text{s.t.} \quad \lambda_c > \lambda_{\text{min}},
\end{equation}
where $\mathcal{A}_x$ is the set of occurring actions in video $x$ with $T_x$ frames, and $\lambda_{\text{min}}$ denotes a pre-set minimum length over all actions. This ensures a minimum difference between estimated lengths by summing composing $\lambda_c$ and the actual length $T_x$ over the video set. \cref{eq:len} can be solved with constrained optimization by linear approximation (COBYLA)~\cite{powell1994direct}. The explicit modeling of lengths is necessary to avoid producing unreasonably long action segments.

The third component, the \emph{visual model}, provides the probability of a feature sequence $\bm{x}$ being generated by the given action labels $\bm{c}$. There are multiple ways to model the frame likelihood. For example, 
\cite{richard2018neuralnetwork} follows Bayes' theorem and estimates $p(x_t|c_n)$ by considering:
\begin{equation}
 p(x_t|c_n) \propto \frac{p(c_n|x_t)}{p(c_n)}.
\end{equation}
The prior $p(c_n)$ can be estimated either empirically, based on the fraction of frames with label $c_n$~\cite{richard2018action} or simply as a uniform distribution~\cite{li2021action}. The posterior $p(c_n|x_t)$ is then approximated by the output of an action classification network supervised by the action annotations.

An alternative way to model the frame likelihood is with a Gaussian mixture model (GMM). In the GMM, the likelihood for a video frame $x_t$ given action class $c_n$ is defined as: 
\begin{equation}\label{eq:likelihood}
 p(x_t|c_n) = \mathcal{N}(x_t; \mu_n, \Sigma_n),
\end{equation}
where $\mu_n$ and $\Sigma_n$ are the action class mean and covariance respectively. In practice, GMMs are preferred for cases where no action annotations are available~\cite{kukleva2019unsupervised,vidalmata2021joint}.

\textbf{Viterbi.} The MAP solution for the HMM described in \cref{eq:hmm} can be solved efficiently with the Viterbi algorithm~\cite{viterbi1967error}. Viterbi uses dynamic programming to find the most likely sequence of states with a given action order. %
Considering the case where a uniform length model is applied, \cref{eq:hmm} yields the following:
\begin{equation}
 (\hat{N}, \hat{\bm{c}}) = \argmax_{N,\bm{c}} \prod_{n=1}^{N-1} p(c_{n+1}|c_n) \cdot \prod_{t=1}^T p(x_t|c_n).
\end{equation}
This can be simplified by denoting the labeling sequence of $T$ frames as $\bm{\pi}$, \ie 
\begin{equation}
 \hat{\bm{\pi}} = \argmax_{\bm{\pi}}\prod_{t=1}^T p(x_t|\pi_t)\cdot p(\pi_t|\pi_{t-1}).
\end{equation}
One can then define the probability $Q_{t,\hat{\pi}_t}$ of the most probable state sequence at time $t$,
\begin{align}
 Q_{1,\pi_1} &= p(x_1|\pi_1)\cdot p(\pi_1) \quad \text{and}\\
 Q_{t,\hat{\pi}_t} &=\max_{\pi_t}\left( p(x_t|\pi_t)\cdot p(\pi_t|\pi_{t-1})\cdot Q_{t-1, \hat{\pi}_{t-1}} \right). \label{eq:vtk}
\end{align}
The Viterbi path $\hat{\bm{\pi}}$ can be retrieved by traversing the stored best $\hat{\pi}_t$ from each timestamp in \cref{eq:vtk}. The overall complexity of this implementation is $\mathcal{O}(T\times |N|^2)$. 

\textbf{Re-estimation.} \cite{kukleva2019unsupervised,li2021action} have stated that the aforementioned HMM model can be updated iteratively. %
As a first step, one initializes the above three HMM components with naive observations. Second, the Viterbi decoding is applied to infer the MAP label sequence. The decoded labels can again be applied to refine the feature inputs to the HMM components. %
These steps can be repeated until convergence.

\textbf{Inference.} To reduce the computational complexity of Viterbi,~\cite{souri2021fifa} proposed FIFA. %
Instead of dynamic programming,~\cite{souri2021fifa} defines a differentiable energy function to approximate the probabilities of possible segment alignments. %
Their inference process reformulates the maximization of the sequence posterior by minimizing the proposed energy function. Given the transcript $c_{1:N}$, the aim is to find the lengths $\ell_{1:N}$ correspondingly, \ie,
\begin{equation}
 \hat{\ell}_{1:N} =\argmin_{\ell_{1:N}}\underbrace{ -\log p(\ell_{1:N}|x_{1:T}, c_{1:N}) }_{E(\ell_{1:N})}.%
\end{equation}
The objective energy function $E(\ell_{1:N})$ can be further decomposed as
\begin{align}
 E(\ell_{1:N}) &= -\log\left(\prod_{t=1}^T p(\alpha(t)|x_t)\cdot\prod_{n=1}^N p(\ell_n|c_n)\right) \nonumber\\
 &=\underbrace{\sum_{t=1}^T -\log p(\alpha(t)|x_t)}_{E_o} + \underbrace{\sum_{n=1}^N -\log p(\ell_n|c_n)}_{E_\ell},\label{eq:energy}
\end{align}
where $p(\alpha(t)) = p(y_t|t;c_{1:N},\ell_{1:N})$, $\alpha(t)$ is the mapping of time $t$ to action label given the segment-wise labeling, and $c_{1:N}$ is sampled from the training set. 

Two further approximations are used for the two terms in~\cref{eq:energy}. First is a differentiable mask $M\in \mathbb{R}^{N\times T}$ with a parametric plateau function $f$~\cite{moltisanti2019action}:
\begin{align}
 M[n,t] &= f(t|\lambda_n^c, \lambda_n^w, \lambda^s) \nonumber \\
 & = \frac{1}{(e^{\lambda^s(t-\lambda_n^c-\lambda_n^w)}+1)(e^{\lambda^s(-t+\lambda_n^c-\lambda_n^w)}+1)},
\end{align}
where $\lambda^c, \lambda^w$ are the center and lengths of a plateau computed from $\ell_{1:N}$ and $\lambda^s$ is a fixed sharpness parameter. Hence, the first term $E_o$ is approximated as:
\begin{equation}
 E_o^* = \sum_{t=1}^T \sum_{n=1}^N M[n,t]\cdot P[n,t]
\end{equation}
where $P[n,t]= -\log p(c_n|x_t)$ is the negative log probabilities. Secondly, for $E_\ell$, $c_n$ is replaced with the expected length value $\lambda_{c_n}^\ell$ based on a Laplace distribution assumption:
\begin{equation}
 E_L^* = \frac{1}{Z} \sum_{n=1}^N |\ell_n - \lambda_{c_n}^\ell |,
\end{equation}
where $Z$ is the constant normalization factor. The original energy function is finally expressed as a weighted aggregation of two approximation terms:
\begin{equation}
 E^*(\ell_{1:N}) = E_o^* (\ell_{1:N}) + \beta E_\ell^*(\ell_{1:N}),
\end{equation}
where $\beta$ is a coefficient. FIFA can boost the inference speed up to 5$\times$ while maintaining a comparable performance score.

\textbf{Generalized Mallows Model (gMM).}
A gMM models distributions over orderings or permutations. Given a set of videos belonging to the same activity, Sener~\etal~\cite{sener2018unsupervised} proposed using a gMM to model the sequential structures of actions for action segmentation. They assume that a canonical sequence ordering $\bm{\sigma}$ is shared in these videos and treats action ordering $\bm{\pi}$ as a permutation of $\bm{\sigma}$. Such a model offers flexibility for missing steps and deviations. A gMM represents permutations as a vector of inversion counts $\bm{v}=[v_1,\cdots, v_{K-1}]$, where $K$ is the number of elements, \ie, actions, in the ordering, and $v_k$ denotes the total number of elements from $(k+1,\cdots, K)$ that rank before $k$ in the ordering $\bm{\pi}$. With the distance between two orderings defined as $d(\bm{\pi}, \bm{\sigma})=\sum_k \rho_k v_k$, the probability of observing $\bm{v}$ is as follows:
\begin{equation}
 P_{\text{gMM}}(\bm{v}|\bm{\rho}) = \frac{e^{-\sum_{k}\rho_k v_k}}{\psi_k(\bm{\rho})} = \prod_k^{K-1}\frac{e^{-\rho_k v_k}}{\psi_k(\rho_k)},
\end{equation}
where $\bm{\rho}=[\rho_1,\cdots, \rho_{K-1}]$ is a set of dispersion parameters and $\psi_k (\rho_k)$ is the normalization function. The prior for each $\rho_k$ is the conjugate:
\begin{equation}
 P(\rho_k|v_{k,0},v_0) \propto e^{-\rho_kv_{k,0}-
\log(\psi_k(\rho_k))v_0},
\end{equation}
A common prior $\rho_0$ is used for each $k$, such that
\begin{equation}
 v_{k,0} = \frac{1}{e^{\rho_0}} -\frac{K-k+1}{e^{(K-k+1)\rho_0}-1}.
\end{equation}
Given an action ordering $\bm{\pi}$, generating frame-wise label assignment $\bm{z}$ requires a bag of action labels $\bm{a}$. $\bm{a}$ is %
modeled as a multinomial parameterized by $\bm{\theta}$. The parameter $\bm{\theta}$ is assumed to follow a Dirichlet prior with hyperparameter $\theta_0$.

The gMM model in~\cite{sener2018unsupervised} aims to find the (latent) set of orderings for the entire video collection, \ie infer the posterior $P(\bm{z},\bm{\rho} | \bm{F},\theta_0,\rho_0, v_0)$, 
where $\bm{F}$ is the frame features, $\bm{z}$ is the temporal action segmentation output and $\theta_0,\rho_0, v_0$ are hyperparameters. 
Similar to HMMs, the above model can also be trained in two stages, where discriminative feature clustering %
and sequential modeling are performed in an alternating fashion.

\subsubsection{Over-Segmentation}
Local continuity is an inherent attribute of procedural actions, meaning an action should be locally continuous and only change at its actual boundary. This has motivated researchers to refine the resulting segments %
at the boundaries to improve performance.

\definecolor{action2}{RGB}{130,179,102}
\definecolor{action1}{RGB}{150,115,166}
\definecolor{action3}{RGB}{150,115,166}

\begin{figure}[!tb]
\centering
    \subfigure[One-hot]{\label{fig:one-hot}
    \hspace{-1em}
        \includegraphics{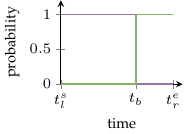}
    }
    \hspace{-1em}
    \subfigure[Linear~\cite{ding2018weakly}]{\label{fig:fixed}
        \includegraphics{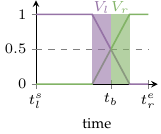}
    }
    \hspace{-1em}
    \subfigure[ABS~\cite{ding2022leveraging}]{\label{fig:abs}
        \includegraphics{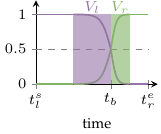}
    }

\caption{Action probability assignment for frames around an action boundary as a function of time. 
    Let $t_b$ denote the estimated boundary between the left action in $[t_l^s,t_b)$ and the right action $[t_b,t_r^e)$. The colour-shaded segments denote the boundary vicinities $V_l$ and $V_r$. (a) One-hot labels adopt a step function with hard action labels assignments. (b) Linear~\cite{ding2018weakly} linearly mixes the action probabilities. (c) ABS~\cite{ding2022leveraging} uses a sigmoid function with a decay proportional to the action duration.
    }\label{fig:bs}
\end{figure}

\textbf{Boundary Refinement.} Wang~\etal~\cite{wangboundary}, motivated to reduce boundary ambiguity and over-segmentation, %
proposed a refinement module for MS-TCN~\cite{farha2019ms}. Their module uses a novel Local Barrier Pooling operation to smooth boundary predictions with confident ones in the later stages of MS-TCN. Separately,~\cite{ishikawa2021alleviating} proposed supplementing segmentation outputs with detected boundary detections from a separate network structure. %
 
\textbf{Gaussian Smoothing.} %
Smoothing with a Gaussian kernel~\cite{ishikawa2021alleviating,ding2021temporal,du2022fast} promotes the continuity of actions in a narrow local temporal window and is highly effective in %
raising segmentation metrics. While \cite{ishikawa2021alleviating,ding2021temporal} directly apply smoothing on the frame-wise action probabilities, Du~\etal~\cite{du2022fast} apply it along the temporal dimension of sequential similarity scores between consecutive frames %
for more robust boundaries.

\textbf{Boundary Smoothing.} From a different perspective, some research suggests that soft action boundaries %
may improve the TAS performance
~\cite{ding2018weakly,ding2022leveraging} over the use of hard transitions. \cref{fig:bs} compares variants of soft action boundaries, %
including linear~\cite{ding2018weakly} %
andsigmoidal~\cite{ding2022leveraging} transitions. %

\section{Level of Supervision}\label{sec:levelofsupervision} 
This section comprehensively compiles modern TAS techniques and categorizes them based on their level of supervision. Section~\ref{sec:full} covers the fully supervised approaches, Section~\ref{sec:weak} discusses the weakly-supervised approaches, Section~\ref{sec:unsup} explores the unsupervised approaches, and finally, Section~\ref{sec:semi} summarizes the semi-supervised approaches.

\subsection{Fully-Supervised Approaches}~\label{sec:full}
\begin{table*}[!htb]
\centering
\caption{Performance of supervised TAS methods on GTEA and Breakfast. \textbf{Repr.} corresponds to approaches targeting learning better feature representations. \textbf{TCN} lists methods built on {T}emporal {C}onvolutional {N}etworks. \textbf{TF} uses {t}rans{f}ormer as the backbone. \textbf{Refine} group aims to improve the performance of existing backbones. 
}\label{tab:supMethods} 
\begin{threeparttable} 
\begin{tabular}{|c|lc|l|ccccc|ccccc|} 
\cline{2-14}
\multicolumn{1}{c|}{} & \multicolumn{1}{l}{\multirow{2}{*}{\textbf{Method}}} & \multirow{2}{*}{\textbf{Year}} & \multicolumn{1}{c|}{\multirow{2}{*}{\textbf{Input/Feature}}} & \multicolumn{5}{c|}{ \textbf{GTEA}} & \multicolumn{5}{c|}{ \textbf{Breakfast}} \\\cline{5-14}
\multicolumn{1}{c|}{} & & & & \multicolumn{3}{c}{F1@\{10, 25, 50\}} & Edit & MoF & \multicolumn{3}{c}{F1@\{10, 25, 50\}} & Edit & MoF \\\cline{2-14}\noalign{\vspace{0.65ex}} 
\hline

\parbox[t]{3mm}{\multirow{8}{*}{\rotatebox[origin=c]{90}{\textbf{Repr.}}}} & \cite{singh2016multi}~Bi-LSTM & 2016 & RGB + flow & 66.5 & 59.0 & 43.6 & - & 55.5 & - & - & - & - & - \\
 & \cite{lea2016segmental}~ST-CNN & 2016 & RGB + motion* & 58.7 & 54.4 & 41.9 & - & 60.6 & - & - & - & - & - \\
 & \cite{mac2019learning}~LCDC & 2019 & RGB + motion* & 52.4 & - & - & 45.4 & 55.3 & - & - & - & - & - \\
 & \cite{gammulle2019coupled}~Coupled GAN & 2019 & RGB + flow & 80.1 & 77.9 & 69.1 & 72.8 & 78.5 & - & - & - & - & - \\
 & \cite{sener2020temporal}~TempAgg & 2020 & I3D & - & - & - & - & - & 59.2 & 53.9 & 39.5 & 54.5 & 64.5 \\
 & \cite{ahn2021refining}~HASR + MS-TCN~\cite{farha2019ms} & 2021 & I3D & 90.0 & 88.1 & 74.8 & 85.6 & 77.5 & \textbf{73.2} & \textbf{68.1} & \textbf{54.0} &  71.0  & \textbf{69.0} \\
 & \cite{ishihara2022mcfm}~MCFM & 2022 & pose + I3D & 91.8 & 91.2 & 80.8 & 88.0 & 80.5 & - & - & - & - & - \\
 & \cite{yi2021asformer}~Br-Prompt + ASFormer~\cite{li2022bridge} & 2022 & image + text & \textbf{94.1} & \textbf{92.0} & \textbf{83.0} & \textbf{91.6} & \textbf{81.2} & - & - & - & - & - \\
\hline\hline

\parbox[t]{3mm}{\multirow{12}{*}{\rotatebox[origin=c]{90}{\textbf{TCN}}}} & \cite{lea2017temporal} ED-TCN & 2017 & LCDC~\cite{mac2019learning} & 75.4 & - & - & 72.8 & 65.3 & - & - & - & - & - \\ 
 & \cite{lea2017temporal} ED-TCN & 2017 & IDT + FV~\cite{kuehne2016end} & - & - & - & - & - & - & - & - & - & 43.3 \\ 
 & \cite{lea2017temporal} ED-TCN & 2017 & spatial-CNN~\cite{lea2016segmental} & 72.2 & 69.3 & 56.0 & - & 64.0 & - & - & - & - & - \\ 
 & \cite{ding2017tricornet} TricorNet & 2017 & spatial-CNN~\cite{lea2016segmental} & 76.0 & 71.1 & 59.2 & - & 64.8 & - & - & - & - & - \\ 
 & \cite{lei2018temporal} TDRN & 2018 & spatial-CNN~\cite{lea2016segmental} & 79.2 & 74.4 & 62.7 & 74.1 & 70.1 & - & - & - & - & - \\ 
 & \cite{farha2019ms} MS-TCN & 2019 & IDT & - & - & - & - & - & 58.2 & 52.9 & 40.8 & 61.4 & 65.1 \\ 
 & \cite{farha2019ms} MS-TCN & 2019 & I3D (FT$\dagger$) & 87.5 & 85.4 & 74.6 & 81.4 & 79.2 & - & - & - & - & - \\ 
 & \cite{farha2019ms} MS-TCN & 2019 & I3D & 85.8 & 83.4 & 69.8 & 79.0 & 76.3 & 52.6 & 48.1 & 37.9 & 61.7 & 66.3 \\ 
 & \cite{li2020ms} MS-TCN++ & 2020 & I3D & 88.8 & 85.7 & 76.0 & 83.5 & 80.1 & 64.1 & 58.6 & 45.9 & 65.6 & 67.6 \\
 & \cite{zhang2019frontal} RPGaussian & 2019 & I3D & 88.5 & 86.8 & 74.6 & 84.0 & 78.5 & 62.0 & 56.0 & 43.7 & 63.5 & 64.2 \\
 & \cite{wang2020gated} GatedR & 2020 & I3D & 89.1 & 87.5 & 72.8 & 83.5 & 76.7 & 71.1 & 65.7 & 53.6 & \textbf{70.6} & 67.7\\
 & \cite{singhania2021coarse} C2F-TCN & 2021 & I3D & \textbf{90.3} & \textbf{88.8} & \textbf{77.7} & \textbf{86.4} & \textbf{80.8} & \textbf{72.2} & \textbf{68.7} & \textbf{57.6} & 69.6 & \textbf{76.0} \\
\hline\hline

\parbox[t]{3mm}{\multirow{4}{*}{\rotatebox[origin=c]{90}{\textbf{TF}}}} & \cite{yi2021asformer} ASFormer & 2021 & I3D & 90.1 & 88.8 & 79.2 & 84.6 & 79.7 & 76.0 & 70.6 & 57.4 & 75.0 & 73.5\\
 & \cite{AZIERE2022104567}TCTr & 2022 & I3D & 91.3 & 90.1 & 80.0 & 87.9 & \textbf{81.1} & 76.6 & 71.1 & 58.5 & 76.1 & \textbf{77.5} \\
 & \cite{du2022dilated} FAMMSDTN & 2022 & I3D & 91.6 & 90.9 & 80.9 & 88.3 & 80.7 & 78.5 & \textbf{72.9} & \textbf{60.2} & \textbf{77.5} & \textbf{74.8}\\ 
 & \cite{behrmann2022unified} UVAST & 2022 & I3D & \textbf{92.7} & \textbf{91.3} & \textbf{81.0}& \textbf{92.1} & 80.2 & \textbf{76.9} & 71.5 & 58.0 & 77.1 & 69.7 \\ 
\hline \hline
\parbox[t]{3mm}{\multirow{15}{*}{\rotatebox[origin=c]{90}{\textbf{Refine}}}} & \cite{chen2020actiona} MTDA + MS-TCN   & 2020 & I3D & 90.5  & 88.4 & 76.2 & 85.8 & 80.0  & 74.2 & 68.6 &  56.5  & 73.6 &  71.0 \\
 & \cite{chen2020action} SSTDA + MS-TCN  & 2020 & I3D & 90.0 &  89.1  & 78.0 &  86.2  & 79.8 &  75.0  &  69.1  & 55.2 & 73.7  & 70.2 \\
 & \cite{huang2020improving} GTRM + MS-TCN** & 2020 & I3D & - & - & - & - & - & 57.5 & 54.0 & 43.3 & 58.7 & 65.0 \\
 & \cite{wangboundary} BCN + MS-TCN & 2020 & I3D & 88.5 & 87.1 & 77.3 & 84.4 &  79.8  & 68.7 & 65.5 & 55.0 & 66.2 & 70.4 \\
 & \cite{ishikawa2021alleviating} ASRF + MS-TCN & 2020 & I3D & 89.4 & 87.8 &  79.8  & 83.7 & 77.3 & 74.3 & 68.9 & 56.1 & 72.4 & 67.6\\ 
 & \cite{gao2021global2local} G2L + MS-TCN & 2021 & I3D & 89.9 & 87.3 & 75.8 & 84.6 & 78.5 & 74.9 & 69.0 & 55.2 & 73.3 & 70.7 \\ 
 & \cite{souri2021fifa} FIFA + MS-TCN & 2021 & I3D & - & - & - & - & - & 75.5 & 70.2 & 54.8 & 78.5 & 68.6\\
 & \cite{souri2021fifa} FIFA + UVAST~\cite{behrmann2022unified} & 2022 & I3D & 82.9 & 79.4 & 64.7 & 90.5 & 69.8 & 76.9 & 71.5 & 58.0 & 77.1 & 69.7\\
 & \cite{xu2022dont} DTL + MS-TCN & 2022 & I3D & - & - & - & - & - & 73.0 & 67.7 & 54.4 & 71.6 & 72.3 \\
 & \cite{park2022maximization} DPRN + MS-TCN & 2022 & I3D & 92.9 & 92.0 & 82.9 & 90.9 & 82.0 & 75.6 & 70.5 & 57.6 & 75.1 & 71.7\\
 & \cite{souri2021fifa} FIFA + ASFormer~\cite{yi2021asformer} & 2021 & I3D & 90.4 & 88.6 & 78.1 & 86.2 & 78.9 & 76.8 & 71.4 & 58.9 & 75.6 & 73.7\\
 & \cite{ijcai2022p115} UARL + ASFormer & 2022 & I3D & 92.7 & 91.5 & 82.8 & 88.1 & 79.6 & 65.2 & 59.4 & 47.4 & 
 66.2 & 67.8 \\
 & \cite{xu2022dont} DTL + ASFormer & 2022 & I3D & - & - & - & - & - & 78.8 & 74.5 & 62.9 & 77.7 & 75.8\\
 & \cite{kim2022stacked} SEDT + ASFormer & 2022 & I3D & \textbf{93.7} & \textbf{92.4} & 84.0 & \textbf{91.3} & 81.3 & - & - & - & - & - \\
 & \cite{liu2023diffusion} DiffAct + ASFormer & 2023 & I3D & 92.5 & 91.5 & \textbf{84.7} & 89.6 & \textbf{82.2} & \textbf{80.3} & \textbf{75.9} & \textbf{64.6} & \textbf{78.4} & \textbf{76.4} \\
\hline 
\end{tabular}
\begin{tablenotes}
 \item[*] Motion images are computed by taking the difference between frames across a 2 second window. 
 \item[**] The improvements are computed based on the authors' implementation of MS-TCN.  
 \item[$\dagger$] FT denotes fine-tuning.
\end{tablenotes}

\end{threeparttable}
\end{table*}
Under full supervision, each frame of every sequence is labelled. Like action recognition, modern action segmentation techniques rely heavily on deep learning. Some pre-deep learning approaches classified actions in a temporal sliding window~\cite{karaman2014fast,rohrbach2012database,cheng2014temporal} and subsequently obtained the final segmentation through post-processing. Cheng~\etal~\cite{cheng2014temporal} used a Bayesian non-parametric language model to reason on the dependencies between actions. Conversely, \cite{fathi2011understanding,fathi2013modeling} modelled the actions as a change in the state of objects and approached the segmentation problem as finding change points. Another line of methods predicted the most probable sequence of actions using stochastic context-free grammars to capture the temporal structure of actions~\cite{vo2014stochastic,pirsiavash2014parsing}. Kuehne~\etal~\cite{kuehne2014language,kuehne2016end} combined the grammars with a set of HMMs, which are used to model the coarse action units. Richard~\etal~\cite{richard2016temporal} introduced a mapping of visual cues to action probability, aided by a language model on the action sequence and a length model on the segment duration. 
 
This section introduces deep-learning based approaches for fully-supervised TAS, highlighting different aspects of representation learning, network architectures and iterative refinement. ~\cref{tab:supMethods} compares most methods' performance on Breakfast and GTEA. 

\subsubsection{Representation Learning}~\label{sec:suppRepresentation}
Initial works combined deep features with temporal models. For example, ST-CNN~\cite{lea2016segmental} uses a CNN to capture spatiotemporal feature relations along with a semi-Markov model. To compute visual representations, Bi-LSTM~\cite{singh2016multi} splits videos into snippets and passes them through a multi-stream (appearance and motion) network similar to ~\cite{simonyan2014two}. The features are then fed into a bi-directional LSTM to predict action labels. 

Follow-up works focused on improving the representations for fine-grained actions. For example, LCDC~\cite{mac2019learning} embeds fine-grained motions with locally consistent deformable convolutions to replace optical flow. Coupled-GAN~\cite{gammulle2019coupled} incorporates two generative adversarial networks, one for RGB images and one for auxiliary data (depth or optical flow), to capture the progression of actions. TempAgg~\cite{sener2020temporal} facilitates multi-granular temporal aggregation that relates recent observations to long-range ones with attention. This network can be used for TAS by naively classifying long-range information-aggregated snippets. While the reported performance is solely based on snippet scores, incorporating a sequence model is expected to yield further enhancements. 
 
Recent approaches focused on enhancing the effectiveness of current backbones by leveraging improved representations~\cite{ahn2021refining} and incorporating multi-modal features~\cite{ishihara2022mcfm,li2022bridge}. HASR~\cite{ahn2021refining} first extracts a segment-level representation for each segment based on the frames and then extracts a video-level representation based on the segment-level representations. Br-Prompt~\cite{li2022bridge} introduces a framework for feature learning based on prompts. They jointly trained video and text encoders at the frame level, using a vision transformer (ViT)~\cite{dosovitskiy2020image} as the video encoder. The jointly learned features are applied to segmentation in ASFormer by~\cite{yi2021asformer}. 

\subsubsection{Architectures}~\label{sec:architecture}
\textbf{Temporal Convolutional Networks (TCNs)} capture temporal patterns with a series of feed-forward convolutional layers. Compared to previous works~\cite{lea2016segmental,singh2016multi}, TCNs are convenient in that they implicitly capture action durations, pairwise transitions, and long-term dependencies, all within the architecture directly. Lea~\etal~\cite{lea2017temporal} are the first to introduce TCNs for temporal action segmentation with an encoder-decoder (ED-TCN) architecture using 1D temporal convolutional and deconvolutional kernels. A follow-up work, TricorNet~\cite{ding2017tricornet} presents a hybrid temporal convolutional and recurrent network by replacing the decoder in ED-TCN with a bi-directional LSTM. However, the recurrence in the decoder incurred large computation costs. TDRN~\cite{lei2018temporal} builds upon ED-TCN by substituting the temporal convolutions with deformable temporal convolutions and adding a residual stream to the encoder-decoder model. The residual stream processes videos at \emph{full} temporal resolution, while the other stream captures temporal context at varying scales. 

The above-mentioned approaches~\cite{lea2016segmental,lea2017temporal,ding2017tricornet} are considered at \emph{`full'} resolution because the segmentation outputs match in frame-rate compared to the original input video. However, the input video is often downsampled to a few frames per second on the encoding side. This type of pre-processing may cause the loss of fine-grained details. In contrast, Farha and Gall~\cite{farha2019ms} proposed a hierarchical multi-stage temporal convolutional network (MS-TCN) that encodes full-resolution video. Each stage of MS-TCN comprises multiple temporal convolutional layers and outputs an initial prediction that is iteratively refined by subsequent stages. Within each stage, the (full) temporal resolution is preserved by progressively dilated convolutions. MS-TCN significantly improves segmentation performance compared to earlier methods~\cite{lea2017temporal,lei2018temporal} by a large margin and reduces over-segmentation errors. The follow-up MS-TCN++~\cite{li2020ms} introduces a dual dilated layer and incorporated parameter sharing in the refinement stages. 

Several follow-up works proposed improvements to the MS-TCN architecture. RPGaussian~\cite{zhang2019frontal} integrates a bilinear pooling module into TCNs by substituting the final $1\times1$ convolution layer in the initial stage of MS-TCN for efficient feature fusion. GatedR~\cite{wang2020gated} adds a gated refinement network to adaptively correct errors from preceding stages. It incorporates a multi-stage sequence-level refinement loss to rectify errors from previous predictions.

C2F-TCN ~\cite{singhania2021coarse} is an encoder-decoder framework with the motivation to tackle over-segmentation. This model incorporates a coarse-to-fine ensemble of decoding layers, resulting in less fragmented segments. This work also introduces a multi-resolution feature-level augmentation strategy and a complex activity loss, improving segmentation performance.

\textbf{Transformers} have emerged as the recent architecture of choice for many vision tasks, including temporal action segmentation. The ASFormer~\cite{yi2021asformer} uses one encoder and multiple decoders. The encoder incorporates dilated temporal convolution and a self-attention layer to initialize the segmentation. The decoders leverage cross-attention to gather information from the encoder and iteratively refine the segmentation from previous blocks. UVAST~\cite{behrmann2022unified} uses a similar encoder structure but changes the decoding into an auto-regressive framework to predict transcripts. Removing the frame-wise prediction allowed UVAST to outperform previous methods on the segment metrics of F1 and Edit Score. To simultaneously address the over-segmentation of TCNs and Transformers requiring substantial amounts of training data, TCTr~\cite{AZIERE2022104567} proposes a hybrid approach. It consists of one convolution stage and four Transformer encoding stages to balance complexity and performance. FAMMSDTN~\cite{du2022dilated} proposes using multi-layer dilated Transformers to capture local and global temporal relationships across different time spans in videos. 

\subsubsection{Segmentation Refinement} 
Multiple works have focused on enhancing existing segmentation backbones by integrating new modules or losses or refining the outputs through post-processing. SSTDA~\cite{chen2020action} argues that spatiotemporal variations of human actions from different videos, referred to as \emph{different domains}, hinder supervised segmentation performance. To mitigate this issue, it proposes to use two self-supervised auxiliary tasks. The first task predicts the domain of unaltered frame-wise feature vectors, while the second task predicts domain labels for a shuffled sequence of segments from both the source and target domains. This form of self-supervision applied to MS-TCN significantly improved performance without requiring additional labeled data. DTL~\cite{xu2022dont} introduces a temporal logic loss that measures the consistency between the output and imposed temporal constraints. GTRM~\cite{huang2020improving} refines segmentations with a graph convolutional network (GCNs). Instead of applying manually defined receptive fields, G2L~\cite{gao2021global2local} introduces a search scheme to identify effective combinations of receptive fields for existing segmentation models. A recent work, DiffAct~\cite{liu2023diffusion} using an ASFormer~\cite{yi2021asformer} backbone, presents state-of-the-art performance using denoising diffusion models. The segmentation outputs are iteratively generated from random noise conditioning on the input video features. Additionally, DiffAct proposes a masking strategy that jointly uses position, boundary, and relation priors of human actions to enhance the segmentation results.

Several post-processing approaches target improving the action boundaries. BCN~\cite{wangboundary} introduces a module for MS-TCN that features a pooling operator to smooth noisy low-confidence boundary predictions with clean, confident ones. ASRF~\cite{ishikawa2021alleviating} refines boundaries with a complementary network branch that directly regresses boundary locations. This approach is model-independent and applicable to any temporal segmentation output. DPRN~\cite{park2022maximization} proposes a divide-and-conquer approach to maximize frame-wise classification accuracy and reduce over-segmentation errors. UARL~\cite{ijcai2022p115} estimates the uncertainty stemming from ambiguous boundaries by Monte-Carlo sampling, while SEDT~\cite{kim2022stacked} introduces a strategy to smooth annotations around boundaries into soft labels.

\subsection{Weakly-Supervised Approaches}\label{sec:weak} 
Weakly supervised techniques aim to minimize the reliance on dense frame-level supervision. Four types of weak supervision are explored in action segmentation: transcripts, action sets, timestamps and text such as narrations. Transcripts and action sets are ordered and unordered lists of actions, respectively, without associated frame information. Timestamps are action labels at specified time-frames. 
 
The performance of these approaches on Breakfast and 50Salads are compared in Table \ref{tab:weaksupMethods}. Using transcripts outperforms methods using action sets, and Timestamps-based approaches outperform all others, indicating higher levels of supervision generally lead to better performance. 
 
\subsubsection{Action Transcripts} 
A \emph{transcript} is a sequential list of actions that occur in a video. This form of supervision offers a notable advantage in terms of cost reduction for video annotation, as it eliminates the need for dense frame-by-frame labels. Methods that learn from transcripts are either iterative two-stage or single-stage solutions.

\textbf{Iterative two-stage solutions} start with an initial estimate of frame-wise labels based on the provided transcript label and progressively improve the previous predictions through iterative refinements. HTK~\cite{kuehne2017weakly} extends their supervised action segmentation approach~\cite{kuehne2016end} to a weakly supervised setting, using HMMs to represent the actions and GMMs to model the observations. The algorithm initializes the video segments uniformly and iteratively refines them based on the provided transcripts. Richard~\etal~\cite{richard2017weakly,kuehne2018hybrid} enhance HTK by replacing the GMMs with RNNs and introduce the latent sub-actions to capture fine-grained motions within the same action. ISBA~\cite{ding2018weakly} starts with a uniform partition of the video according to the provided transcript and adjusts the action boundary gradually with a soft labeling scheme. TASL~\cite{lu2021weakly} by iteratively aligning training videos based on the transcripts. 

\textbf{Single-stage solutions} argue that the two-step approaches~\cite{kuehne2017weakly,richard2017weakly} are initialization-sensitive and may not always converge. ECTC~\cite{huang2016connectionist} extends the temporal classification from~\cite{graves2006connectionist} to align transcripts with video frames while imposing consistency restrictions. It enforces frame-wise similarities to ensure consistent action alignments. 

NN-Viterbi~\cite{richard2018neuralnetwork} uses Viterbi decoding to generate pseudo-labels from transcripts to train their framework composed of visual, context and length models. It presents a significant performance advancement compared to previous methods. D$^3$TW~\cite{chang2019d3tw} uses a differentiable alignment loss to model positive and negative transcripts discriminatively. Similarly, CDFL~\cite{li2019weakly}, which builds upon NN-Viterbi, uses discriminative transcript modeling. However, unlike D$^3$TW, CDFL generates valid and invalid candidates using a segmentation graph; invalid candidates violate the transcripts. CDFL also introduces a new loss based on energy differences between the valid and invalid candidates using a recursive estimation of each candidate's segmentation energy. Although NN-Viterbi or CDFL offers stronger performance than preceding approaches, their training process entails higher computational costs due to the Viterbi decoding. Souri~\etal~\cite{souri2021fast} emphasized this extensive training time and proposed MuCon, a sequence-to-sequence framework with comparable performance while significantly reducing the training and inference time. MuCon features two network branches, one predicting frame-wise actions and the other predicting transcripts with durations; the two branches are linked with a mutual consistency loss. 

DP-DTW~\cite{chang2021learning} addresses weakly-supervised segmentation by training class-specific discriminative action prototypes. It represents videos by concatenating prototypes based on transcripts and enhances inter-class distinction between prototypes through discriminative losses.

\begin{table}[!tb]
\centering
\caption{Performance of weakly supervised methods on Breakfast and 50Salads. \textbf{Tr} indicates action {tr}anscripts, \textbf{T} is for iterative {t}wo-stage solutions, while \textbf{S} is for {s}ingle-stage. Action {sets} is denoted by \textbf{Set}. \textbf{TS} is for {t}ime{s}tamp supervision.   
}\label{tab:weaksupMethods}
 
\adjustbox{max width=\linewidth}{%
\begin{tabular} {|c|lc|l|ccc|c|} 
\cline{2-8}
\multicolumn{1}{c|}{} & \multicolumn{1}{l}{\multirow{2}{*}{\textbf{Method}}} & \multirow{2}{*}{\textbf{Year}} & \multicolumn{1}{c|}{\multirow{2}{*}{\textbf{Feature}}} & \multicolumn{3}{c|}{ \textbf{Breakfast}} & \textbf{50Salads}\\ \cline{5-8}
\multicolumn{1}{c|}{} & & & & MoF & IoU & IoD & MoF \\ \cline{2-8}\noalign{\vspace{0.65ex}}\hline

{\multirow{4}{*}{\rotatebox[origin=c]{90}{\textbf{Tr + T}}}} & \cite{kuehne2017weakly}~HTK & 2017 & IDT + FV & 25.9 & - & - & 24.7 \\ 
 & \cite{richard2017weakly}~HMM/RNN & 2017 & IDT + FV & 33.3 & - & - & \textbf{45.5} \\ 
 & \cite{ding2018weakly}~ISBA & 2018 & IDT + FV & 38.4 & 24.2 & \textbf{40.6} & -\\ %
 & \cite{lu2021weakly}~TASL & 2021 & IDT + FV & \textbf{49.9} & \textbf{36.6} & 34.3 & - \\
\hline\hline

{\multirow{6}{*}{\rotatebox[origin=c]{90}{\textbf{Tr + S}}}} & \cite{huang2016connectionist}~ECTC & 2016 & IDT + FV & 27.7 & - & - & - \\ 
 & \cite{richard2018neuralnetwork}~NN-Viterbi & 2018 & IDT + FV & 42.9 & 32.2 & 29.1 & 49.4 \\ 
 & \cite{chang2019d3tw}~D$^3$TW & 2019 & IDT + FV & 45.7 & - & - & -\\ 
 & \cite{li2019weakly}~CDFL & 2019 & IDT + FV & 50.2 & 33.7 & \textbf{45.4} & \textbf{54.7}\\ 
 & \cite{souri2021fast}~MuCon & 2019 & IDT + FV & 48.5 & - & - & -\\ 
 & \cite{chang2021learning}~DP-DTW & 2021 & IDT + FV & \textbf{50.8} & \textbf{35.6} & 45.1 & -\\ 
\hline\hline

{\multirow{6}{*}{\rotatebox[origin=c]{90}{\textbf{Set}}}}&\cite{richard2018action} ActionSet~\cite{richard2018action} & 2018& IDT + FV& 23.3 & - & - & -\\
 & \cite{fayyaz2020sct}~SCT & 2020 & IDT + FV & 26.6 & - & - & -\\
 & \cite{fayyaz2020sct}~SCT & 2020 & I3D & 30.4 & - & - & -\\
 & \cite{li2020set}~SCV & 2020 & IDT + FV & 30.2 & - & - & -\\
 & \cite{li2021anchor}~ACV & 2021 & IDT + FV &  33.4 & - & - & -\\
 & \cite{lu2022set}~POC & 2022 & I3D & \textbf{42.4} & \textbf{33.5} & - & -\\
\hline\hline

{\multirow{4}{*}{\rotatebox[origin=c]{90}{\textbf{TS}}}} & \cite{li2021temporal} Timestamps & 2021 & I3D & \textbf{64.1} & - & - & {75.6}\\ 
 & \cite{rahaman2022generalized}~EM-TSS & 2022 & I3D & 63.7 & - & - &  75.9 \\ 
 & \cite{khan2022timestamp}~GCN-TSS & 2022 & I3D & 61.4 & - & - & 75.1\\ 
 & \cite{souri2022robust}~RAS-TSS & 2022 & I3D & 62.9 & - & - & \textbf{79.3} \\ 
\hline 
 
\end{tabular}}

\end{table}

\subsubsection{Action Sets} 
Action sets are a unique set of the actions in a given video; they are a weaker form of supervision than action transcripts because they lack the action ordering and the frequency, \ie how often an action occurs within an action. Such labels often appear as meta-tags on video-sharing platforms. 
 
Richard~\etal~\cite{richard2018action} are the first to propose a weak segmentation model based on action sets. Building upon~\cite{richard2016temporal}, their framework leverages a context, length and action model to find the sequence of actions that maximizes the overall likelihood for a given video, which can be solved via Viterbi. To limit the search space, this framework generates candidate transcripts using a context-free grammar; the problem is effectively transformed into the action transcript setting with multiple transcripts. However, it should be noted that candidate transcripts may not fully cover all possible action sequences, limiting segmentation accuracy. SCT~\cite{fayyaz2020sct} learns a segmentation network directly from the action sets using a set prediction loss. Their model initially segments videos into regions, followed by action probabilities and temporal lengths estimation using one branch. A second branch generates frame-wise action predictions; the two branches are linked with a consistency loss between frame-wise and region predictions. 

SCV~\cite{li2020set} uses a set-constrained Viterbi to generate pseudo ground truths and an $n$-pair loss to minimize the cosine distance between pairs of training videos sharing action classes in their respective action sets. To ensure that all actions in the set are considered in the frame-wise pseudo-ground truth, SCV adds a greedy post-processing to assign the missing action label to the segments such that the assignment's posterior probability is minimally decreased. Building on SCV, ACV~\cite{li2021anchor} introduces a differentiable approximation for end-to-end training and removes the need for post-processing. 
 
POC~\cite{lu2022set} highlights that in multiple videos, action pairs tend to demonstrate a consistent temporal order. To leverage this insight, POC introduces a pairwise order consistency loss that penalizes disagreements in ordering between the extracted templates and the segmentation model outputs.

\subsubsection{Timestamp Supervision} 
Timestamp supervision provides action labels for a sparse set of frames instead of the dense video sequence. Some methods~\cite{li2021temporal,khan2022timestamp} place constraints on this set, \ie one frame from each action (equivalent to an augmented transcript in which each action is matched to a single frame), while others~\cite{rahaman2022generalized} are more relaxed and allow for arbitrarily sampled frames. The general strategy of timestamps is to generate pseudo frame-wise labels for supervision, then refine them iteratively. Notably, timestamp methods perform comparably with fully supervised approaches, making it a compelling direction for further exploration.
 
Li~\etal~\cite{li2021temporal} introduced a novel approach to predict frame-wise labels by detecting action transitions. Their model incorporates a confidence loss to encourage class probabilities to decrease monotonically with respect to distance from the timestamp. EM-TSS~\cite{rahaman2022generalized} uses Expectation-Maximization to infer the missing frame labels from labeled timestamps. The expectation step trains the network to estimate frame-wise labels while the maximization maximizes timestamp segment likelihoods to estimate action boundaries. One key advantage of the EM formulation is that it accommodates arbitrarily sampled frames and can handle either missing segments and or segments with multiple timestamps. This makes it more flexible, annotation-wise, compared to~\cite{li2021temporal}. EM-TSS also shows that selecting the boundary frames as timestamps for each action segment impairs performance compared to using random or middle frames, highlighting the ambiguity of labels at the boundaries.

GCN-TSS~\cite{khan2022timestamp} proposes a Graph Neural Network (GNN) framework where frame features are treated as nodes, and the edges between consecutive frames are weighted based on their feature affinity. The GNN is trained to propagate labels from a small number of labeled nodes to the remaining unlabeled nodes. These works assume that every action instance is annotated with a timestamp, meaning no actions are missed by the annotators. RAS-TSS~\cite{souri2022robust} relaxes this assumption and allows for missing annotations for some action classes. It also expands the segment boundaries beyond the timestamps in both directions rather than solely detecting action changes between two consecutive timestamps to minimize the number of frames with unknown annotations. 
 
\begin{table*}[!htb]
\centering
\caption{Performance of unsupervised methods evaluated on Breakfast. \textbf{Two Stage} indicates the two-stage learning, \textbf{Joint} indicates joint learning, \textbf{Video} denotes single video-based clustering methods and finally the methods works with the \textbf{G}lobal video corpus. \textbf{A} correspond to activity level evaluation while \textbf{V} the video level. Their \textbf{Temporal Model}s are defined and corresponding flexibility for  \textbf{Deviations}, \textbf{Missing} steps and \textbf{Repetitions} in orderings are compared. 
}\label{tab:unsupMethods} 
\adjustbox{max width=\textwidth}{
\begin{threeparttable}

\begin{tabular}{|c|lc|l|ccc|l|ccc|}\cline{2-11}
\multicolumn{1}{c|}{} & \multicolumn{1}{l}{\textbf{Method}} & \textbf{Year} & \multicolumn{1}{c|}{\textbf{Input/Feature}} & \textbf{F1(A)}& \textbf{MoF(A)} & \textbf{MoF(V)} & \multicolumn{1}{c|}{\textbf{Temporal Model}} & \textbf{Deviations} & \textbf{Missing} & \textbf{Repetitions}\\ \cline{2-11}\noalign{\vspace{0.65ex}} \hline

\parbox[t]{3mm}{\multirow{7}{*}{\rotatebox[origin=c]{90}{\textbf{Two Stage}}}} & \cite{sener2018unsupervised}~Mallows& 2018 & IDT + FV & - & 34.6 & - & Mallows model~\cite{fligner1986distance}& \boldcheckmark & \boldcheckmark & - \\ 
&\cite{goel2019learning} Prism& 2019 & IDT + FV & - & 33.5 & - & hierarchical Bayesian model & - & - & \boldcheckmark\\ 
&\cite{kukleva2019unsupervised}~CTE& 2019 & IDT + FV & 26.4 & 41.8 & - & temporal cluster order & - & \boldcheckmark & -\\ 
&\cite{vidalmata2021joint}~JVT & 2021& IDT + FV & 29.9 & 48.1 & 52.2 & temporal cluster order & - & \boldcheckmark & -\\
& \cite{li2021action}~ASAL & 2021 & IDT + FV & 37.9 & 52.5 & - & HMM & - & \boldcheckmark & -\\
&\cite{wang2022sscap}~CAP & 2021 & SpeedNet~\cite{benaim2020speednet} & \textbf{39.2} & 51.1 & - & temporal cluster order & - & \boldcheckmark & \boldcheckmark\\
&\cite{lin2023taec}~TAEC & 2023 & IDT + FV & 33.6& 50.3& 62.6 & global cluster assignment& \boldcheckmark & \boldcheckmark & -\\
\hline\hline 

\parbox[t]{3mm}{\multirow{3}{*}{\rotatebox[origin=c]{90}{\textbf{Joint}}}} 
&\cite{swetha2021unsupervised}~UDE& 2021& I3D & 31.9 & 47.4 & \textbf{74.6}& temporal cluster order & - & \boldcheckmark & -\\
&\cite{kumar2022unsupervised}~TOT& 2021& IDT + FV & 31.0 & 47.5 & - & temporal optimal transport & - & \boldcheckmark & - \\ 
&\cite{tran2023permutation}~UFSA & 2023 & IDT + FV & 38.0 & 52.1 & - & temporal optimal transport & \boldcheckmark & \boldcheckmark & - \\
\hline\hline

\parbox[t]{3mm}{\multirow{4}{*}{\rotatebox[origin=c]{90}{\textbf{Video}}}} & \cite{aakur2019perceptual}~LSTM+AL & 2019& CNN~\cite{simonyan2014very} & - & - & 42.9 & - & - & - & -\\ 
&\cite{sarfraz2021temporally}~TW-FINCH & 2021 & IDT + FV& -& -& 62.7 & - & - & - & -\\
&\cite{du2022fast}~ABD & 2022 & IDT + FV & -& - & 64.0 & - & - & - & - \\ 
&\cite{bueno2023leveraging}~TSA & 2023 & IDT + FV & -& - & 65.1 & - & - & - & - \\ 
\hline\hline

\parbox[t]{3mm}{\multirow{2}{*}{\rotatebox[origin=c]{90}{\textbf{G}}}}&\cite{ding2021temporal}~CAD & 2021 & IDT + FV & -&49.5& - & temporal cluster order& - & \boldcheckmark & -\\
&\cite{ding2021temporal}~CAD & 2021 & I3D & -&\textbf{53.1} & - & temporal cluster order & - & \boldcheckmark & -\\
\hline

\end{tabular}
\end{threeparttable} 
}

\end{table*}

\subsubsection{Narrations \& Subtitles} 
Frequently, videos are accompanied by publicly available text data in the forms of scripts, subtitles or narrations. Text data is commonly used for video via text alignment~\cite{malmaud2015s,bojanowski2015weakly} and step localization~\cite{alayrac2016unsupervised,zhukov2019cross}. Multi-modal learning from video data has gained significant attention recently, as it enables the learning of rich embedding spaces that facilitate zero-shot retrieval without the need for explicit labels. While most of these works~\cite{radford2021learning,zhang2021vinvl} focus on vision-language learning, some~\cite{shvetsova2022everything} also incorporate audio information. One major drawback of using textual data is the assumption that they are temporally well aligned to the visual context for \emph{all} videos. Unfortunately, this assumption does not always hold and could be completely void. 

Sener~\etal~\cite{sener2015unsupervised} proposed a hybrid generative model that combines visual and language cues for video segmentation. This model generates visual vocabularies from the frame-wise object proposals of a collection of videos depicting the same activity and textual vocabularies from the narrative text. Each frame is then represented using a binary histogram of visual and textual words. Then, a generative beta process mixture model~\cite{fox2014joint} is utilized to detect actions shared across multiple videos. Fried~\etal~\cite{fried2020learning} modeled activities based on a canonical ordering of actions that are assumed to be given during inference. Contrary to transcript-based approaches, which assume the transcript is provided per video, Fried~\etal~\cite{fried2020learning} assume one transcript per activity class. They applied a semi-Markov model to capture segment duration, location, order, and features. 

More recent works explore the text information accompanying instructional videos and temporally align narrations to corresponding video segments~\cite{han2022temporal}. 

\begin{table*}[!ht]
\centering
\caption{Performance of semi-supervised methods evaluated on Breakfast, 50Salads and GTEA with varying ratios of labeled data (D\%). Abbreviated names are {f}eature {l}earning (\textbf{FL}) that learns a new set of inputs in a self-supervised manner, {t}est {d}ata (\textbf{TD}) is used for feature learning, and {f}eature {e}nsembling (\textbf{FE}) is employed to boost the performance. {C}omplex {a}ctivity (\textbf{CA}) denotes the usage of video-level labels for training, which is only applicable to Breakfast. 
}\label{tab:semisupMethods}
\adjustbox{max width=\textwidth}{
\begin{tabular}{|c|lc|c|c|c|c|c|ccccc|ccccc|ccccc|}\hline %
\multicolumn{1}{|c|}{\multirow{2}{*}{\textbf{D\%}}} & \multicolumn{1}{l}{\multirow{2}{*}{\textbf{Method}}} & \multirow{2}{*}{\textbf{Year}} & \multirow{2}{*}{\textbf{FL}} & \multirow{2}{*}{\textbf{TD}} & \multirow{2}{*}{\textbf{FE}} & \multirow{2}{*}{\textbf{Backbone}} & \multicolumn{6}{c|}{\textbf{Breakfast}} & \multicolumn{5}{c|}{\textbf{50Salads}} & \multicolumn{5}{c|}{\textbf{GTEA}}\\\cline{8-23}
 & & & & & & & \textbf{CA} & \multicolumn{3}{c}{F1@\{10, 25, 50\}} & Edit & Acc & \multicolumn{3}{c}{F1@\{10, 25, 50\}} & Edit & Acc & \multicolumn{3}{c}{F1@\{10, 25, 50\}} & Edit & Acc\\\hline\hline

\multirow{5}{*}{5} & \cite{ding2022leveraging} SemiTAS & 2022 & - & - & - & MS-TCN~\cite{farha2019ms} & - & 44.5 & 35.3 & 26.5 & 45.9 & 38.1 & 37.4 & 32.3 & 25.5 & 32.9 & 52.3 & 59.8 & 53.6 & 39.0 & 55.7 & 55.8 \\
 & \cite{ding2022leveraging} SemiTAS & 2022 & - & - & - & MS-TCN~\cite{farha2019ms} & \boldcheckmark & 56.6 & 49.3 & \textbf{35.8} & \textbf{59.4} & 56.6 & - & - & - & - & - & - & - & - & - & - \\
 & \cite{singhania2022iterative} ICC & 2022 & \boldcheckmark & \boldcheckmark & \boldcheckmark & ED-TCN~\cite{lea2017temporal} & - & - & - & - & - & - & 39.3 & 34.4 & 21.6 & 32.7 & 46.4 & - & - & - & - & -\\
 & \cite{singhania2022iterative} ICC & 2022 & - & - & \boldcheckmark & C2F-TCN~\cite{singhania2022iterative} & - & - & - & - & - & - & 42.6 & 37.5 & 25.3 & 35.2 & 53.4 & - & - & - & - & -\\
 & \cite{singhania2022iterative} ICC & 2022 & \boldcheckmark & \boldcheckmark & \boldcheckmark & C2F-TCN~\cite{singhania2022iterative} & \boldcheckmark & \textbf{60.2} & \textbf{53.5} & 35.6 & 56.6 & \textbf{65.3} & \textbf{52.9} & \textbf{49.0} & \textbf{36.6} & \textbf{45.6} & \textbf{61.3} & \textbf{77.9} & \textbf{71.6} & \textbf{54.6} & \textbf{71.4} & \textbf{68.2} \\\hline\hline

\multirow{2}{*}{10} & \cite{ding2022leveraging}~SemiTAS & 2022 & - & - & - & MS-TCN~\cite{farha2019ms} & - & 56.9 & 51.3 & 39.0 & 57.7 & 49.5 & 47.3 & 42.7 & 31.8 & 43.6 & 58.0 & 71.5 & 66.0 & 52.9 & 67.2 & 62.6 \\
 & \cite{singhania2022iterative}~ICC & 2022 & \boldcheckmark & \boldcheckmark & \boldcheckmark & C2F-TCN~\cite{singhania2022iterative} & \boldcheckmark & \textbf{64.6} & \textbf{59.0} & \textbf{42.2} & \textbf{61.9} & \textbf{68.8} & \textbf{67.3} & \textbf{64.9} & \textbf{49.2} & \textbf{56.9} & \textbf{68.6} & \textbf{83.7} & \textbf{81.9} & \textbf{66.6} & \textbf{76.4} & \textbf{73.3}\\ \hline
\end{tabular}}
\end{table*} 

\subsection{Unsupervised Approaches}\label{sec:unsup}
Unsupervised TAS approaches operate without any labels. However, due to the task definition and nature of the proposed learning strategies, many works inherently require activity labels~\cite{sener2015unsupervised, sener2018unsupervised, kukleva2019unsupervised} since they're applied to videos of each activity independently. Table~\ref{tab:unsupMethods} compares unsupervised methods on Breakfast. The general strategies of unsupervised approaches either follow a two-step iterative process that alternates between representation learning and frame-wise clustering or simultaneous representation learning and clustering. Additional works focus on detecting boundary changes within single stand-alone videos. A final category leverages the same amount of activity-level information but utilizes the global video corpus. 

\subsubsection{Two-Stage Learning}
Sener and Yao~\cite{sener2018unsupervised} were the first to present an unsupervised segmentation method using only visual inputs. Their approach alternates between discriminative learning of action appearance and generative modeling of the action sequence with a gMM~\cite{fligner1986distance}. The gMM is a distribution over permutations and can capture ordering variations and missing steps but cannot handle repeated actions. Prism~\cite{goel2019learning} presents a hierarchical generative Bayesian model that can accommodate repeated actions. However, this model assumes all videos adhere to the same underlying ordering.
 
Deep-learning based frameworks CTE~\cite{kukleva2019unsupervised} and JVT~\cite{vidalmata2021joint} advance frame-wise representation learning with continuous temporal~\cite{kukleva2019unsupervised} and visual-temporal~\cite{vidalmata2021joint} embeddings. Clustering is then done in the embedding space before finding the videos' ordering with Viterbi decoding. Both works assume a fixed sequential order determined by the average timestamp within each cluster. While this assumption accommodates missing steps, it can not handle deviations or repetitions. TAEC~\cite{lin2023taec} presents a two-stage pipeline comprising a sequence-to-sequence temporal embedding network and a cross-video global clustering approach allowing deviations from a fixed action order.

Building upon previous unsupervised works~\cite{kukleva2019unsupervised,vidalmata2021joint}, SS-CAP~\cite{wang2022sscap} computes the video order by representing the multi-occurrence of actions using co-occurrence relations. Furthermore, SS-CAP~ conducts a comparative analysis of different self-supervised approaches to demonstrate performance improvements using various feature-learning techniques. 
ASAL~\cite{li2021action} presents an efficient solution for self-supervised feature embedding learning by temporarily shuffling the predicted action segments and classifying the resulting action sequences as valid or invalid. 
 
\subsubsection{Joint Representation Learning and Clustering}
UDE~\cite{swetha2021unsupervised} is the first to jointly learn the embedding and clustering. It combines visual and positional encodings and uses contrastive learning for clustering in a latent space. TOT~\cite{kumar2022unsupervised} uses a combination of temporal optimal transport to maintain the temporal order of actions and a temporal coherence loss to preserve affinity across adjacent frames. A recent transformer-based method, UFSA~\cite{tran2023permutation}, learns action prototypes and frame embedding simultaneously and uses a temporal optimal transport module to align them and produce pseudo labels. UFSA integrates the encoder from ASFormer~\cite{yi2021asformer} and the decoder from UVAST~\cite{behrmann2022unified}, enabling the utilization of segment-level cues. Notably, including a segment-level prediction module and a frame-to-segment alignment module injects flexibility in action order.

\subsubsection{Single Video Clustering} 
These works have a different focus from the previous three groups in that they target detecting boundaries without considering action dynamics. This concept follows early unsupervised segmentation methods doing change-point detection with temporal sliding windows in music~\cite{harchaoui2009regularized}, financial data~\cite{xuan2007modeling}, and event clustering~\cite{zelnik2001event}. Interestingly, these methods, either evaluated at the video level or activity level, outperform unsupervised temporal segmentation techniques. This can partly be attributed to the scale of existing datasets being too small to effectively emphasize the importance of sequential modeling~\cite{fathi2011learning} or predominant activities adhering to a fixed ordering~\cite{kuehne2014language}.

LSTM+AL~\cite{aakur2019perceptual} introduces a self-supervised method for detecting action boundaries in a single pass over videos. It predicts the feature of the next frame; action boundaries are determined based on discrepancies between predicted and observed features. TW-FINCH~\cite{sarfraz2021temporally} captures the spatiotemporal similarities among frames and used a temporally weighted hierarchical clustering algorithm to group video frames. This approach does not require training since it directly operates on pre-computed features. Similarly, ABD~\cite{du2022fast} identifies action boundaries by detecting abrupt change points along the similarity chain calculated between consecutive features. TSA~\cite{bueno2023leveraging} focuses on learning action representations for individual videos by training a shallow network using a triplet loss and a novel triplet selection strategy. The resulting learned representations can be processed with a generic clustering algorithm to obtain segmentation outputs.

\subsubsection{Global Video Corpus} 
Unlike the previous unsupervised works that work with a single collection of videos from the same activity, CAD~\cite{ding2021temporal} leverages the activity labels to learn on the entire video corpus for TAS. This framework proposes learning frame representations based on their similarity to the latent action prototypes. It assumes that the complex activity label can be inferred using aggregated action prototype affinities across the whole video sequence. 
 
\subsection{Semi-Supervised Approaches}\label{sec:semi}
In contrast to weak supervision, which requires annotations for \emph{every} training video, semi-supervised learning requires dense frame-wise labels for a \emph{subset} of videos. Existing approaches either leverage the unlabeled set for better representation or mine priors from them. Detailed performance comparisons are given in~\cref{tab:semisupMethods}.

SemiTAS~\cite{ding2022leveraging} demonstrates that even a small subset of densely annotated videos provides more informative cues than timestamp supervision applied to the entire dataset. SemiTAS shows that such supervision offers valuable action-level priors to guide learning on unlabeled videos. It introduces two novel loss functions for semi-supervised TAS: the action affinity loss and the action continuity loss. Specifically, the affinity loss enforces action composition and distribution priors by minimizing the Kullback-Leibler divergence between the closest pairs of labeled and unlabeled videos. ICC~\cite{singhania2022iterative} learns a new set of feature representations through unsupervised contrastive learning. These features are subsequently used for training a classifier within the semi-supervised setting. The network's predictions serve as pseudo-labels for supervising the unlabeled videos. Remarkably, ICC achieves comparable performance to fully-supervised counterparts with only 40\% of labeled videos. 

\section{Conclusions and Outlook}\label{sec:conclusion}
This survey provided an overview of the temporal action segmentation techniques followed by a thorough evaluation of existing works. The substantial body of literature highlights the growing interest and attention in the task of action segmentation. However, despite the rapid progress, there are still numerous unexplored areas that we invite the research community to explore and discuss some aspects below.
 
\textbf{Input Features.} The majority works on TAS typically take visual feature vectors, either hand-crafted (IDT)~\cite{wang2013action} or extracted from an off-the-shelf CNN backbone (I3D)~\cite{carreira2017quo}, as input for each frame. Using pre-computed inputs is conventional practice for several other tasks as well, including temporal action localization~\cite{shou2016temporal}, action anticipation~\cite{abu2018will,miech2019leveraging}, as it greatly reduces the computational demands. Nonetheless, as pointed out by~\cite{choi2019can,huang2018makes}, pre-computed characteristics tend to favour static cues, \eg, scene components, within frames. To the best of our knowledge, no empirical research has compared utilizing pre-computed features to training TAS models from raw images end-to-end due to the high demands in terms of training efficiency and GPU memory requirements.
 
\textbf{Segment-Level Modeling.} 
As mentioned in \cref{subsec:seqmodel}, the majority of current methods for the sequential modeling of actions are iterative and independent of feature learning. Sequential modeling approaches are especially common for post-processing and refining per-frame outputs. Exploring how to add sequence-based losses, such as edit scores that penalize segment-wise mistakes, into the learning process is an interesting but under-explored direction. Segment-level losses readily coincides with the first interpretation of the TAS task (\cref{eq:taskseg}) while the majority of existing techniques follow frame-wise prediction (\cref{eq:taskframe}). We recommend a greater emphasis on solving the task at the segment level and anticipate that it can greatly reduce the current issues with oversegmentation.
 
\textbf{Forms of Supervision.} 
Procedural video sequences feature enormous temporal redundancy in the supervisory signals due to the significant similarity between successive video frames of the same motion. Such redundancy is been proven by the comparable performance of using timestamp supervision~\cite{li2021temporal} vs. fully-supervised setting~\cite{li2021temporal,rahaman2022generalized}. While annotating only action timestamps significantly reduces the effort required, it still demands a vigilant annotator to review each video diligently, ensuring no activities are overlooked. Briefly explored in~\cite{rahaman2022generalized}, how to handle missing actions in annotations is an open direction for TAS.

An additional component of supervision to consider is the inherent uncertainty of the action boundaries, as actions occurring in time are frequently not as distinct as an object in space. According to~\cite{ding2022leveraging}, these uncertainties in action boundaries can have a significant impact on model performance. It is, therefore, worthwhile to investigate how to define/label action boundaries.

\textbf{Downstream tasks.} Temporal segments can be used as inputs to downstream tasks. For instance,~\cite{soran2015generating} segments video streams in order to send alerts regarding missed actions. While~\cite{rivoir2019unsupervised} uses segmentation as a preliminary job for estimating the remaining time in lengthy surgery videos. Similarly, many approaches in action anticipation use segmentation techniques to represent prior observations with action labels~\cite{abu2018will,Ke_2019_CVPR,gammulle2019forecasting}. This is because such labels contain high-level semantic information, which is preferred over visual characteristics for anticipation tasks~\cite{sener2020temporal}. An intelligent system that has acquired high-level semantic results via a TAS approach can summarize the contents of a movie, \ie, video summarization~\cite{apostolidis2021video}.
 
Moreover, transferring the TAS to an online environment could make these strategies more useful to real-world applications. The initial attempts to achieve this objective were in~\cite{du2022fast,ghoddoosian2022weakly}. Yet, both approaches rely on frame-wise pre-computed features. The online segmentation of videos with end-to-end models could be a future trend.
 
In conclusion, TAS is a promising and rapidly evolving scientific topic with numerous potential real-world applications. In this survey, we present a detailed taxonomy of the problem, a systematic analysis of the fundamental methodologies, and a curated collection of current works classified by levels of supervision. We also highlight the chances and obstacles that lie ahead and hope that this survey will help promote the growth of the community.

\ifCLASSOPTIONcaptionsoff
 \newpage
\fi

\bibliographystyle{IEEEtran}
\bibliography{references_short}

\end{document}